\documentclass[journal]{IEEEtran}

\usepackage{comment}
\usepackage{blindtext}
\usepackage{amsmath, amssymb, amsthm} 
\usepackage{stmaryrd}
\usepackage[dvipsnames]{xcolor}
\usepackage{hyperref}
\usepackage{flushend}
\usepackage{soul}

\newcommand{\eoe}[1]{\overset{!}{#1}}

\usepackage{graphicx}
\usepackage[cm]{fullpage}
\usepackage{natbib}

\usepackage{array}

\begin{document}

\title{
%Representing data structures with sparse activity patterns
%Dimensionality-preserving variable binding for sparse distributed representations 
Variable Binding for Sparse Distributed Representations: Theory and Applications 
}
%\author{E. Paxon Frady$^{1,2}$, Denis Kleyko$^{2,3}$, and Friedrich T. Sommer$^{1,2}$\\ 
%1: NCL, Intel Labs\\
%2: RCTN, UC Berkeley\\
%3: Research Institutes of Sweden}
 
\author{E. Paxon Frady, Denis Kleyko, and Friedrich T. Sommer\\ 
%\thanks{Manuscript received on June 15, 2020. D. Kleyko is supported by the European Union’s Horizon 2020 research and innovation programme under the Marie Skłodowska-Curie Individual Fellowship grant agreement No. 839179. He also acknowledges the support from DARPA's VIP program under Super-HD project.}
\thanks{E. P. Frady and F. T. Sommer are with Neuromorphic Computing Lab, Intel Labs and also with the Redwood Center for Theoretical Neuroscience at the University of California, Berkeley, CA 94720, USA.}% 
\thanks{D. Kleyko is with the Redwood Center for Theoretical Neuroscience at the University of California, Berkeley, CA 94720, USA and also with Intelligent systems Lab at Research Institutes of Sweden, 164 40 Kista, Sweden.
  }% 
}

\maketitle

\begin{abstract}
Symbolic reasoning and neural networks are often considered incompatible approaches in artificial intelligence. Connectionist models known as Vector Symbolic Architectures (VSAs) can potentially bridge this gap by enabling symbolic reasoning with distributed representations \citep{Plate1994, Gayler1998, Kanerva1996}. 
However, classical VSAs and neural networks 
are still incompatible because they represent information differently. VSAs encode symbols by dense pseudo-random vectors, where information is distributed throughout the entire neuron population. Neural networks encode features locally, by the activity of single neurons or small groups of neurons, often forming sparse vectors of neural activation \citep{hinton1986learning}. Following \citet{Rachkovskij2001, Laiho2015}, we explore symbolic reasoning with sparse distributed representations.   

The core operations in VSAs are dyadic operations between vectors to express variable binding and the representation of sets. Thus, algebraic manipulations enable VSAs to represent and process data structures of varying depth in a vector space of fixed dimensionality. Using techniques from compressed sensing, we first show that variable binding between dense vectors in classical VSAs \citep{Gayler1998} is mathematically equivalent to tensor product binding \citep{Smolensky1990} between sparse vectors, an operation which increases dimensionality. This theoretical result implies that dimensionality-preserving binding for general sparse vectors must include a reduction of the tensor matrix into a single sparse vector. 

Two options for sparsity-preserving variable binding are investigated. One binding method for general sparse vectors extends earlier proposals to reduce the tensor product into a vector, such as circular convolution \citep{Plate1994}. The other variable binding method is only defined for sparse block-codes \citep{gripon2011}, block-wise circular convolution  \citep{Laiho2015}. Our experiments reveal that variable binding for block-codes has ideal properties, whereas binding for general sparse vectors also works, but is lossy, similar to previous proposals \citep{Rachkovskij2001}. 
We demonstrate a VSA with sparse block-codes in example applications, cognitive reasoning and classification, and discuss its relevance for neuroscience and neural networks. 

%To understand this gap, and spur ideas for bridging it, we use compressed sensing  \citep{candes2006, donoho2006}. 
%Compressed sensing provides a framework for generating a compressed dense representations with unique one-to-one correspondence to a high-dimensional sparse representation. 
%For each operation defined in vector symbolic models between dense distributed vectors, compressed sensing reveals the operation it induces between the sparse feature vectors. The two essential operations in vector symbolic models are ``addition'' to represent a set, and a type of ``multiplication'' to represent a conjunction or key-value pair, known as the binding operation. Our results show that: 1) Set formation of dense vectors is equivalent to concatenation of the sparse vectors; 2) Binding by component-wise multiplication is equivalent to the tensor product of the sparse vectors, a binding operation proposed in \citet{Smolensky1990}. Interestingly, both operations induced for the sparse vectors do not preserve the dimensionality, i.e., compound objects have representations with higher dimensions than atomic objects.

\end{abstract}


\section{Introduction}
In a traditional computer, the internal representation of data is organized by data structures. A data structure is a collection of data values with their relationships. For example, a simple data structure is a key-value pair, relating a variable name to its assigned value. Particular variables within data structures can be individually accessed for computations. Data structures are the backbones for computation, and needed for organizing, storing, managing and manipulating information in computers. 

For many tasks that brains have to solve, for instance analogical inference in cognitive reasoning tasks and invariant pattern recognition, it is essential to represent knowledge in data structures and to query the components of data structures on the fly. It has been a long-standing debate if, and if so how, brains can represent data structures with neural activity and implement algorithms for their manipulation \citep{fodor1988connectionism}. 

Here, we revisit classical connectionist models \citep{Plate1994, Kanerva1996, Gayler1998} that propose encodings of data structures with distributed representations. Following \citet{Gayler2003}, we will refer to these models as {\it Vector Symbolic Architectures} (VSAs), synonymously their working principles are sometimes summarized as  {\it hyperdimensional computing} \citep{Kanerva2009}.  Typically, VSA models use dense random vectors to represent atomic symbols, such as variable names and feature values. Atomic symbols can be combined into compound symbols that are represented by vectors that have the same dimension. %as atomic symbols and still resemble random vectors. 
The computations with pseudo-random vectors in VSAs rest on the concentration of measure phenomenon \citep{ledoux2001concentration} that random vectors become almost orthogonal in large vector spaces \citep{Frady2018}.

%For example to answer the question ``What is the dollar of Mexico?'' \citep{Kanerva2010}, one can use existing data structures that represent trivia about the US and Mexico, such as their currencies, language, capital, etc. The question is answered by querying the US data structure with ``dollar'' to obtain ``currency'', and then the data structure holding information about Mexico is queried with ``currency'' to obtain ``peso''.

%In invariant pattern recognition, one needs a representation of sensory input where ``shape'' and ``transformation'' can be queried separately. Traditional localist feature representations use vectors with one-hot encodings to describe the presence of a particular feature. 

In neural networks, 
%the representation of information is quite different from VSAs, 
features are encoded locally by the activity of a single or of a few neurons. Also, patterns of neural activity are often sparse, i.e. there are only few nonzero elements \citep{willshaw1969non, Olshausen1996}.  
Connectionists attempted to use such local feature representations in models describing computations in the brain. %and for applications such as invariant pattern recognition.
However, a critical issue emerged with these representations, known as {\it the binding problem} in neuroscience. This problem occurs when a representation requires the encoding of sets of feature conjunctions, 
%It was unclear how such models could form representations of feature conjunctions without confusion, 
for example when representing a red triangle and a blue square \citep{treisman1998}.
Just representing the color and shape features would lose the binding information that the triangle is red, not the square.
One solution proposed for the binding problem is the tensor product representation (TPR) \citep{Smolensky1990}, where a neuron is assigned to each combination of feature conjunctions. However, when expressing hierarchical data structures, the dimensionality of TPRs grows exponentially with hierarchical depth. One proposal to remedy this issue is to form reduced representations of TPRs, so that the resulting representations have the same dimensions as the atomic vectors \citep{Hinton1990, Plate1993}. This has been the inspiration of VSAs, which have proposed various algebraic operations for binding that preserve dimensionality. Building on earlier work on sparse VSA \citep{Rachkovskij2001, Laiho2015}, we investigate the possibility to build binding operations for sparse patterns that preserve dimensionality and sparsity.

% Discussion point?
%There is reason to believe that the currently most successful type of artificial neural networks, deep learning networks, have only limited capabilities to represent and manipulate data structures (CITE). This could explain some of their seemingly fundamental weaknesses, such as the inability to generalize and extrapolate learned knowledge, and the need of large amounts of training data (CITE). 

%In contrast, inspired by functional properties of real neurons, artificial neural networks and TPRs use sparse feature representations -- where a single neuron represents the value of a feature or a particular conjunction, independent from values in other parts of the vector. 

The paper is organized as follows. In section~\ref{sec:bg}, the background for our study is introduced, which covers different flavors of symbolic reasoning, sparse distributed representations, and the basics of compressed sensing. In section~\ref{sec:cs}, compressed sensing is employed to establish the equivalence between the dense representations in classical VSA models and sparse representations. This treatment reveals the operations between sparse vectors that are induced by VSA operations defined on dense vectors. Interestingly, we find that the classical dimensionality-preserving operations in VSAs induce equivalent operations between sparse vectors that do not preserve dimensionality. 
%This theoretical treatment reveals that the dimensionality-preserving binding operation between dense vectors is mathematically equivalent to the TPR of the corresponding sparse vectors. 
Section~\ref{sec:dso} introduces and investigates concrete methods for dimensionality- and sparsity-preserving variable binding. 
%techniques to reduce the TPR to a single sparse vector. 
Known binding methods, such as circular convolution \citep{Plate1994} and vector-derived transformation binding \citep{gosmann2019}, lead to binding operations that are dimensionality- but not sparsity-preserving. We investigate two solutions for sparsity-preserving binding, one for general sparse vectors, and one for the subset of sparse block vectors (block-codes). % and show that it fulfills the requirements of VSAs. 
Section~\ref{sec:apps} demonstrates the most promising solution, a VSA with sparse block-codes in two applications. In  Section~\ref{sec:discu}, we summarize our results and discuss their implications.

\section{Background}
\label{sec:bg}
\subsection{Models for symbolic reasoning}
Many connectionist models for symbolic reasoning with vectors use vector addition (or a thresholded form of it) to express sets of symbols. But the models characteristically deviate in encoding strategies and in their %types of random vectors and the 
operation for binding. TPRs \citep{Smolensky1990} use real-valued localist feature vectors ${\mathbf x},{\mathbf y} \in \mathbb{R}^{N}$ and the outer product ${\mathbf x} \; {\mathbf y}^{\top} \in \mathbb{R}^{N \times N}$ as the binding operation. This form of tensor product binding encodes compound data structures by representations that have higher dimensions than those of atomic symbols. The deeper a hierarchical data structure, the higher the order of the tensor. 
%This rapidly increases the dimensionality of the representation.

Building on Hinton's concept of reduced representations \citep{Hinton1990a}, several VSA models were proposed \citep{Plate1994, Kanerva1996, Gayler1998} in which atomic and composed data structures have the same dimension. These models encode atomic symbols by pseudo-random vectors and the operations for set formation and binding are designed in a way that representations of compound symbols still resemble random vectors. The operations for {\it addition} ($+$) and {\it binding} ($\circ$) are dyadic operations that form a ring-like structure. 
%Addition is usually a regular vector addition or a component-wise thresholded vector addition. 
The desired properties for a binding operation are:
\begin{itemize}
\item[i)] Associative, i.e., $(\mathbf{a} \circ \mathbf{b}) \circ \mathbf{c} = \mathbf{a} \circ (\mathbf{b} \circ \mathbf{c}) = (\mathbf{a} \circ \mathbf{c}) \circ \mathbf{b}$.
% This one is a bit confusing still... I changed M to D_1 because this is not the same M from before. maybe \Phi_i and \Psi_j instead of a^i and b^i
\item[ii)] Distributes over addition, i.e.,\\ $\sum_i^{D_1} \mathbf{a}^i \circ \sum_j^{D_2} \mathbf{b}^j = \sum_{i,j}^{D_1,D_2} \mathbf{c}^{ij}$ with $\mathbf{c}^{ij}= \mathbf{a}^i \circ \mathbf{b}^j$.
\item[iii)] Has an inverse operation to perform unbinding.
\end{itemize}
%VSA models as well as the TPR model have been demonstrated to be able to represent structured knowledge in a way that this knowledge can be used to solve cognitive reasoning problems (see, e.g.~\citep{BuildBrain}). 

{\it Holographic Reduced Representation (HRR)} \citep{Plate1991,Plate1995} was probably the earliest formalized VSA which uses real-valued Gaussian random vectors and circular convolution as the binding operation. 
Circular convolution is the standard convolution operation used in the discrete finite Fourier transform which can be used to produce a vector from two input vectors $\mathbf{x}$ and $\mathbf{y}$:
\noindent
\begin{equation}
    ({\mathbf x} \circ {\mathbf y})_k := (\mathbf{x} \ast \mathbf{y})_k = \sum_{i=1}^N x_{(i-k)_{\mbox{\tiny mod} N}} y_{i} 
    \label{circonv}
\end{equation}
%The addition operation in all the VSA models is just a vector addition, in some cases, like spatter codes, combined with some thresholding or majority rule. 
Other VSA models use binding operations based on projections of the tensor product matrix that only sample the matrix diagonal. %We will call these operation {\it ``local''} because they involve only operations between the pairs of components in the two vectors with the same index.   
%component-wise multiplication of two vectors for binding, the Hadamard product, which is just the main diagonal in the tensor product matrix. 
For example, the {\it Binary Spatter Code (BSC)} \citep{Kanerva1996} uses binary random vectors and binding is the XOR operation between components with the same index. 

In the following, we focus on the {\it Multiply-Add-Permute (MAP) model} \citep{Gayler1998}, which uses bipolar atomic vectors whose components are -1 and 1.  
%To introduce the notation, we will go through the example of representing in a VSA a set of features, each feature assigned to a particular scalar value.
Atomic features or symbols are represented by random vectors of a matrix $\mathbf{\Phi}$, called the {\it codebook}. The columns of $\mathbf{\Phi}$ are normalized i.i.d. random \emph{code vectors}, ${\bf \Phi}_i  \in \{\pm 1\}^N$.
%The binding operation is used to assign values to a feature. In the MAP model, 
The binding operation is the Hadamard product between the two vectors:
\begin{equation}
    {\mathbf x} \circ {\mathbf y} := \mathbf{x} \odot \mathbf{y} = (x_1 y_1, x_2 y_2,..., x_N y_N)^\top
    \label{hadamard}
\end{equation}
%where ${\mathbf{W}_D}$ is the third-order tensor to project the diagonal of the outer-product matrix into a vector.

When the binding involves just a scalar value, the multiplication operation (\ref{hadamard}) relaxes to ordinary vector-scalar multiplication. A feature with a particular value is simply represented by the vector representing the feature, $\mathbf{\Phi}_i$ (which acts like a ``key''),  multiplied with the scalar representing the ``value'' $a_i$: $\mathbf{x} = {\mathbf \Phi}_i a_i$.
  
%represent an analog value of a feature, the value is multiplied with the corresponding hypervector. 
For representing a {\it set of features}, the generic vector addition is used, and the vector representing a set of features with specific values is then given by:   
\begin{equation}
{\bf x} = \mathbf{\Phi} {\bf a}
\label{vsa_summation}
\end{equation}
%We call the vector $\mathbf{a}$ the {\it coefficient vector} of this compound representation, which describes the feature set and the associated feature values. 
Here, the nonzero components of $\mathbf{a}$ represent the values of features contained in the set, the zero-components label the features that are absent in the set.

Although the representation ${\bf x}$ of this set is lossy, a particular feature value can be approximately decoded by forming the inner product with the corresponding ``key'' vector: 
\begin{equation}
a_i \approx  {\bf \Phi}_i^\top \mathbf{x} / N,
\label{eqn:vsa_readout}
\end{equation}
\noindent
where $N$ is the dimension of vectors.
The cross-talk noise in the decoding (\ref{eqn:vsa_readout}) decreases with the square root of  the dimension of the vectors or by increasing the sparseness in $\mathbf{a}$, for analysis of this decoding procedure, see \citep{Frady2018}. 

To represent a set of sets, one cannot simply form a sum of the compound vectors. This is because a feature binding problem occurs, and the set information on the first level is lost. VSAs can solve this issue by combining addition and binding to form a representation of a set of compound objects in which the integrity of individual objects is preserved. This is sometimes called the {\it protected sum} of $L$ objects:
\begin{equation}
\mathbf{s} = \sum_{j}^L \mathbf{\Psi}_j \odot \mathbf{x}^j
\label{prot_sum}
\end{equation}
where $\mathbf{\Psi}_j$ are dense bipolar random vectors that label the different compound objects. %Together they form a codebook of random vectors $\mathbf{\Psi} \in \{-1,1\}^{N \times L}$.
Another method for representing protected sums uses powers of a single random permutation matrix $\mathbf{P}$ \citep{Laiho2015, Frady2018}:
\begin{equation}
\mathbf{s} = \sum_j^L (\mathbf{P})^{(j-1)} \mathbf{x}^j 
\label{eqn:permutprot_sum}
\end{equation}

In general, algebraic manipulation in VSAs yields a noisy representation of the result of a symbolic reasoning procedure. To filter out the result, so-called {\it cleanup memory} is required, which is typically nearest-neighbor search in a content-addressable memory or associative memory \citep{willshaw1969non, palm1980associative, Hopfield1982} storing the codebook(s).    

\subsection{Sparse distributed representations}
The classical VSAs described in the previous section use dense representations, that is, vectors in which most components are nonzero. 
In the context of neuroscience and neural networks for unsupervised learning and synaptic memory, another type of representation has been suggested: {\it sparse representations}. In sparse representations, a large fraction of components are zero, e.g. most neurons are silent. Sparse representations capture essential aspects of receptive field properties seen in neuroscience when encoding sensory inputs, such as natural images or natural sound \citep{Olshausen1996,bell1997}. 

Here, we will investigate how sparse representations can be used in VSAs.
For the cleanup required in VSAs, sparse representations have the advantage that they can be stored more efficiently than dense representations in Hebbian synapses \citep{willshaw1969non, palm1980associative, Tsodyks1988, palmsommer1992information, frady2019}.  However, how the algebraic operations in VSAs can be performed with sparse vectors has only been addressed in a few previous studies
%We investigate whether the VSA operations between dense vectors have corresponding operations for sparse vectors, following earlier work on sparse VSAs
\citep{Rachkovskij2001, Laiho2015}.

A particular type of sparse representation with additional structure has been proposed for symbolic reasoning before: {\it sparse block-codes} \citep{Laiho2015}. In a $K$-sparse block-code, the ratio between active components and total number of components is $K/N$, as usual. But the index set is partitioned into $K$ blocks, each block of size $N/K$, with one active element in each block.
%$N$ the total number of components (neurons) in the vector.
Thus, the activity in each block is maximally sparse, it only contains a single nonzero component\footnote{Note that sparse block-codes differ from sparse block signals \citep{eldar2010block}, in the latter the activity within blocks can be non-sparse but the nonzero blocks is $K'$-sparse, with $K' << K$. The resulting $N$-dimensional vectors have a ratio between active components and total number of components of $K'L/N = K'/K$.}.

The constraint of a sparse block-code reduces the entropy in a code vector significantly, from $\log\left( {N \choose K}\right)$ to $K \log\left(\frac{N}{K}\right)$ bits \citep{SurveyAM17}. At the same time, the block constraint can also be exploited to improve the retrieval in Hebbian associative memory. As a result, the information capacity of associative memories with Hebbian synapses for block-coded sparse vectors is almost the same as for unconstrained sparse vectors \citep{gripon2011,KnoblauchPalm2019}. 
%Here, building on \citet{Laiho2015}, we will leverage sparse block-codes for engineering an efficient binding operation that preserves dimension and sparsity. 
Sparse block-codes also may reflect coding principles observed in the brain, such as competitive mechanisms between sensory neurons representing different features \citep{heeger1992normalization}, as well as orientation hypercolumns seen in the visual system of certain species \citep{hubel1977ferrier}.  

Recent proposals also include sparse phasor-codes for representing information, where the active elements in the population are not binary, but complex-valued with binary magnitudes and arbitrary phases \citep{frady2019}. Such a coding scheme may be relevant for neuroscience, as they can be represented with spikes and spike-timing. VSA architectures have also been demonstrated in the complex domain \citep{Plate2003}, which use dense vectors of unit-magnitude phasors as atomic symbols. Here, we also propose and analyze a variation of the block-code where active entries are phasors.

\subsection{Compressed sensing}
Under certain conditions, there is unique equivalence between sparse and dense vectors that has been investigated under the name \emph{compressed sensing} (CS) \citep{candes2006, donoho2006}. Many types of measurement data, such as images or sounds, have a sparse underlying structure and CS can be used as a compression method, in applications or even for modeling communication in biological brains \citep{hillar2015can}.  For example, if one assumes that the data vectors are $K$-sparse, that is:
\begin{equation}
    \mathbf{a} \in {\cal A}_K := \left\{\mathbf{a} \in I\!\!R^M : ||\mathbf{a}||_0 \leq K\right\}
\end{equation}
with $||.||_0$ the L0-norm.
In CS, the following linear transformation creates a dimensionality-compressed dense vector from the sparse data vector: 
\begin{equation}
%{\bf x} = \mathbf{\Phi} {\bf a},   \;\;\; {\bf y} = \mathbf{\Phi} {\bf b} \  \in I\!\!R^N
%{\bf x} = \mathbf{\Phi} {\bf a},   \;\;\; {\bf y} = \mathbf{\Psi} {\bf b} \  \in I\!\!R^N
\mathbf{x} = \mathbf{\Xi} \mathbf{a}
\label{cs_compression}
\end{equation}
%$\mathbf{\Phi}$ and $\mathbf{\Psi}$ are $N \times M$ random matrices.
where $\mathbf{\Xi}$ is a $N \times M$ random sampling matrix, with $N < M$. % well only when considering CS, but not in VSA
%The vectors ${\bf a}$ and ${\bf b}$ are 
%, where the support is drawn randomly from the $M \choose K$ possible supports and the magnitude of the nonzero vector components is drawn from a Gaussian. 

Due to the distribution of sparse random vectors $\mathbf{a}$, the statistics of the dimensionality-compressed dense vectors $\mathbf{x}$ becomes somewhat non-Gaussian. The data vector can be recovered from the compressed vector $\mathbf{x}$ by solving the following sparse inference problem:
\begin{equation}
\hat{{\bf a}} = \mbox{argmax}_{\mathbf{a}} (\mathbf{x}- \mathbf{\Xi} {\bf a})^2 + \lambda |{\bf a}|_1
\label{cs_recovery}
\end{equation}
The condition for $K$, $N$, $M$ and $\mathbf{\Xi}$, under which the recovery (\ref{cs_recovery}) is possible, forms the cornerstones of compressed sensing \citep{donoho2006,candes2006}.

For CS to work, a necessary condition is that the sampling matrix is injective for the sparse data vectors, i.e. that the intersection between the kernel of the sampling matrix, $\mbox{Ker}(\mathbf{\Xi}) = \{ \mathbf{a}: \mathbf{\Xi} \mathbf{a} = 0\}$, with the set of sparse data vectors, ${\cal A}_K$, is empty: $\mbox{Ker}(\mathbf{\Xi}) \cap {\cal A}_K = \emptyset$. But, this condition does not guarantee that each data vector has a unique dense representation. In other words, the mapping between data vectors and dense representations must also be bijective. To guarantee uniqueness of the dense representation of $K$-sparse vectors, the kernel of the sampling matrix must not contain any $(2K+1)$-sparse vector:
\begin{equation}
    \mbox{Ker}(\mathbf{\Xi}) \cap {\cal A}_{2K+1} = \emptyset
    \label{cs_uniqueness}
\end{equation}
with ${\cal A}_{2K+1}$ being the set of $(2K+1)$-sparse vectors. Intuitively, the condition (\ref{cs_uniqueness}) excludes that any two $K$-sparse data vectors can have the same dense representation: ${\bf a}_1 \not = {\bf a}_2$: $\mathbf{\Xi} {\bf a}_1 - \mathbf{\Xi} {\bf a}_2 = 0$. 

Even with condition (\ref{cs_uniqueness}), it still might not be possible to infer  the sparse data vectors from the dense representations (\ref{cs_recovery}) in the presence of noise. Another common criterion for CS to work is the $s$-restricted isometry property (RIP):
\begin{equation}
(1- \delta_s) ||{\bf a}^s||_2^2 \leq    || \mathbf{\Xi} {\bf a}^s||_2^2 \leq (1+ \delta_s)||{\bf a}^s||_2^2
    \label{rip_cond}
\end{equation}
with the vector ${\bf a}^s$ $s$-sparse, and the RIP constant $\delta_s \in (0,1)$. 
The choice $\delta_{2K+1} =1$ is equivalent to condition (\ref{cs_uniqueness}). With a choice $\delta_{2K+1} = \delta^* <1$, one can impose a more stringent condition that enables the inference, even in the presence of noise. 
%If the compressed representations of two different sparse vectors is identical, a unique reconstruction of the sparse vectors from the compressed representation is impossible.  
The minimal dimension of the compression vector that guarantees (\ref{rip_cond}) is typically linear in $K$ but increases only logarithmically with $M$:
\begin{equation}
    N \geq C \; K \; \log\left(\frac{M}{K}\right)
    \label{cs_cond}
\end{equation}
where $C$ is a constant of order $O(1)$ that depends on $\delta_{2K+1}$. 
%Thus, as long as (\ref{cs_cond}) is fulfilled, the dense vectors can be used as compressed versions of the sparse vectors.
Here, we will use the uniqueness conditions (\ref{cs_uniqueness}) and (\ref{rip_cond}) to assess the equivalence between different models of symbolic reasoning.  

% \subsection{Outline}
% In this report we investigate multiplication or binding operations that are applicable to sparse vectors. The challenge with sparse vectors is that simple local multiplication, like in the Hadamard product, can no longer be applied because with each multiplication, the resulting vectors become sparser. Our investigation will reveal novel insights into relationships between existing models of symbolic reasoning with vectors. 

%and differences of the various coding schemes and use these results for developing a framework of neural representation that can seamlessly cover a problem space ranging from low-level pattern recognition in sensory input to high-level cognitive reasoning. 

\section{Results} 

\subsection{Equivalent representations with sparse vs. dense vectors}
\label{sec:cs}

In this section, we consider a setting where 
sparse and dense symbolic representations can be directly compared. Specifically, we ask what operations between $K$-sparse vectors are induced by the operations in the MAP VSA. To address this question, we map $K$-sparse feature vectors to corresponding dense vectors via (\ref{vsa_summation}). 
The column vectors of the codebook in (\ref{vsa_summation}) correspond to the atomic dense vectors in the VSA.
%while the dense vector resulting from (\ref{vsa_summation}) represent the sets corresponding to the sparse atomic vectors. 
We choose the dimension $N$ and properties of the codebook(s) and sparse random vectors so that the CS condition (\ref{cs_uniqueness}) is fulfilled\footnote{In compressed sensing, the choice of sampling matrices with binary or bipolar random entries is common, e.g,  \citep{amini2011deterministic}.}. 
Thus, each sparse vector
%, for example, an atomic sparse vector selected to represent a symbol, 
has a unique dense representation and vice versa. %Note that an atomic dense vector corresponds to a binary one-hot sparse vector.

\subsubsection{Improved VSA decoding based on CS}

\begin{figure*}[h]
    \centering
    \includegraphics[width=0.9\textwidth]{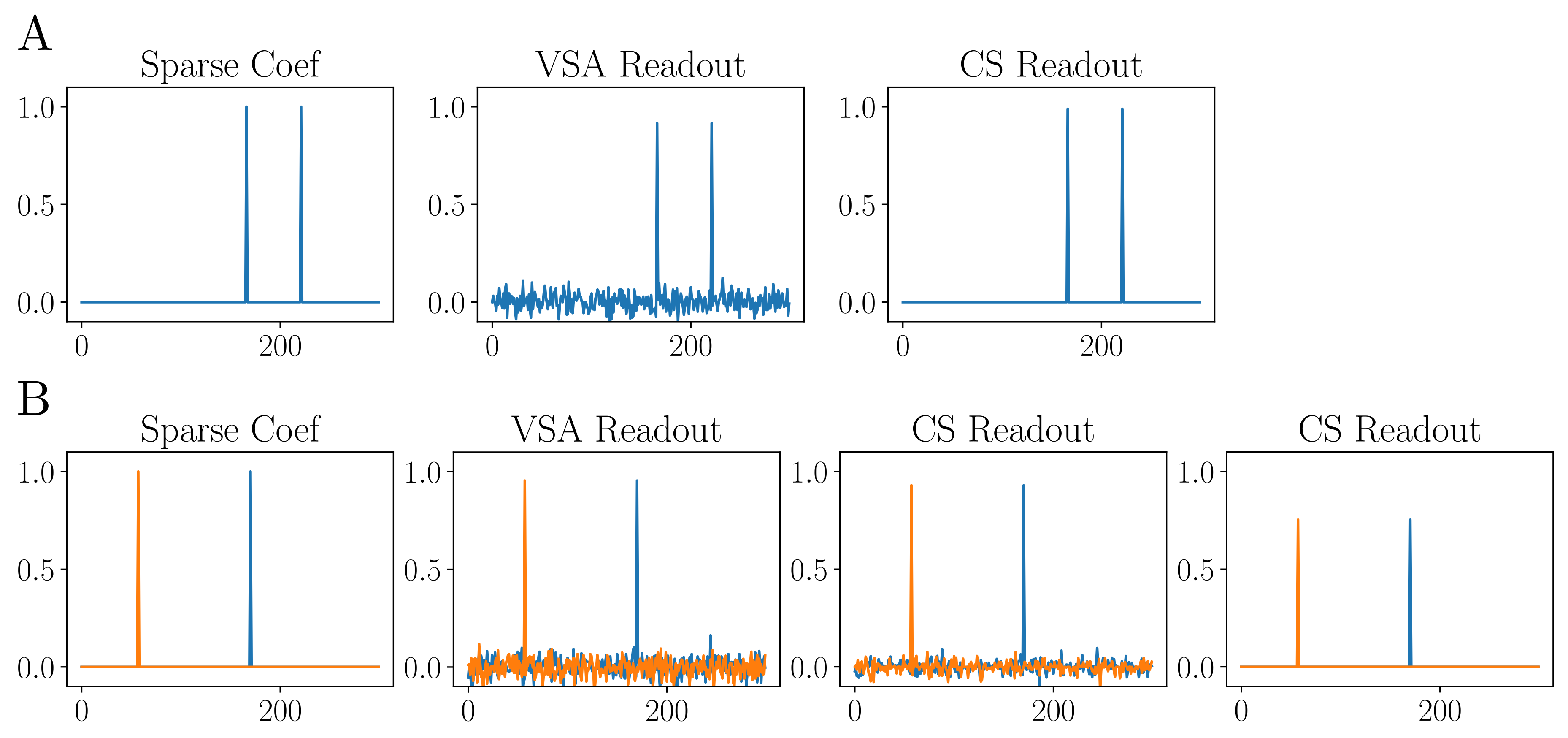}
    \caption{\textbf{Readout of sparse coefficients from dense distributed representation.} A. The sparse coefficients (left) are stored as a dense representation using a random codebook. The coefficients are recovered with standard VSA readout (middle) and with sparse inference (right), which reduces the crosstalk noise. B. Two sparse coefficients are stored as a protected set (left). Readout with sparse inference reduces crosstalk noise, but some noise can remain depending on sparsity penalty.}
    \label{fig:sparse_readout}
\end{figure*}

In our setting, the coefficient vector $\mathbf{a}$ is sparse. The standard decoding method in VSA (\ref{eqn:vsa_readout}) provides a noisy estimate of the sparse vector (Fig.~\ref{fig:sparse_readout}) from the dense representation.  
%Consider the decoding of the key-value pairs in a VSA set representation (\ref{vsa_summation}). 
However, if the sparse vector and the codebook $\mathbf{\Phi}$ in (\ref{vsa_summation}) satisfy the compressed sensing conditions, one can do better: decoding {\`a} la CS (\ref{cs_recovery}) achieves near-perfect accuracy 
%than classical VSA decoding (\ref{eqn:vsa_readout}) 
(Fig.~\ref{fig:sparse_readout}). Note that sparse inference requires that the entire coefficient vector $\mathbf{a}$ is decoded at once, similar to \citet{Ganguli2010a}, while with (\ref{eqn:vsa_readout}) individual values $a_i$ can be decoded separately. 
If the CS condition is violated, sparse inference (\ref{cs_recovery}) abruptly ceases to work, while the VSA decoding with (\ref{eqn:vsa_readout}) gradually degrades, 
%For a theory describing the gradual degradation of decoding accuracy in the case of 1-hot and Gaussian $\mathbf{a}$ vectors, 
see \citet{Frady2018}.

% Second, compression, that is $N<M$, is typical for obtaining efficient dense representations. But, if the dimension of the distributed representations is large,
% %the same or larger than the dimension of the sparse feature vectors,
% then the sparse features are represented along quasi-orthogonal axes in the dense space. Thus, the dense representations are essentially rotated versions of the original sparse representations with all the correlations between sparse features unchanged. The resulting representations are in essence still feature based and not distributed. Conversely, if the random projection is compressive, the sparse features are randomly mixed and through the central limit theorem the components of the distributed codes become more Gaussian and less correlated.

\subsubsection{Variable binding operation}
\label{hadamard_tensor}

%In VSA models the \emph{binding operation} is used for forming representations of conjunctive features. The VSA binding operation creates a new vector for the compound concept that is quasi-orthogonal to the original vectors, as well as to any other conjunction of features. In many VSA models, the binding operation is the Hadamard product between vectors. One proposal about how to represent conjunctions of sparse feature representations involves the tensor product between feature vectors. This results in a representation in which one neuron fires for each combination of features. Next, we show that these binding computations are in fact related.

%The corresponding binding operation between sparse vectors is given by the following {\it Theorem I}:
%s are dimensionality-preserving. For hypervectors with components $\pm 1$ or with complex-valued components on the unit phase circle, the binding operation is defined by component-wise multiplication, specifically, by the Hadamard product of the two dense representations. This 

The Hadamard product between dense vectors turns out to be a function of the tensor product, i.e. the TPR, of the corresponding sparse vectors:
\begin{align}
\begin{split}
 ({\bf x} \odot {\bf y)}_i &= (\mathbf{\Phi} {\bf a})_i  (\mathbf{\Psi} {\bf b})_i = \sum_l \Phi_{il} a_l \sum_k \Psi_{ik} b_k \\
  &= \sum_{lk} \Phi_{il}  \Psi_{ik} a_l  b_k = ((\mathbf{\Phi} \boxdot \mathbf{\Psi}) \ vec( {\bf a} \; {\bf b}^{\top}))_i
%{\bf z}_i:=&({\bf x} \odot {\bf y)}_i = (\mathbf{\Phi} {\bf a})_i  (\mathbf{\Phi} {\bf b})_i = \sum_l \Phi_{il} a_l \sum_k \Phi_{ik} b_k \\
%=& \sum_{lk} \Phi_{il}  \Phi_{ik} a_l  b_k = ((\mathbf{\Phi}^{\odot 2}) \ vec( {\bf a} \otimes {\bf b}))_i %((\mathbf{\Phi} \odot \mathbf{\Phi}) ( {\bf a} \otimes {\bf b}))_i
\label{bind}
\end{split}
\end{align}
%where the RHS expression is the $i$-th component of a matrix-vector product. 
%Equation (\ref{bind}) shows that the Hadamard product of the dense vectors can be expressed as a projection of the outer product of the sparse vectors. 
This linear relationship between the Hadamard product of two vectors and the TPR can be seen as a generalization of the Fourier convolution theorem, see Appendix~\ref{sec:classical_vsa_binding}. 

The reshaping of the structure on the RHS of (\ref{bind}) also shows that there is a relationship to the matrix vector multiplication in CS sampling (\ref{cs_compression}): 
The ravelled tensor product matrix of the sparse vectors becomes a $M^2$-dimensional vector $vec( {\bf a} \; {\bf b}^{\top})$ with $K^2$ nonzero elements. 
%The sparse vector $vec({\bf a} \otimes {\bf b})$ is , with the indices $l$ and $k$, reshaped into a vector of dimension $M^2$. If ${\bf a}$ and ${\bf b}$ are $k$-sparse, $vec({\bf a} \otimes {\bf b})$ is $k^2$-sparse.
Further, $(\mathbf{\Phi} \boxdot \mathbf{\Psi})$ %$(\mathbf{\Phi} \odot \mathbf{\Phi})$ 
is a $N \times M^2$ sampling matrix, formed by pair-wise Hadamard products of vectors in the individual dictionaries $\mathbf{\Phi}$ and $\mathbf{\Psi}$:
\begin{equation}
 (\mathbf{\Phi} \boxdot \mathbf{\Psi})  
:= (\mathbf{\Phi}_1 \odot \mathbf{\Psi}_1, \mathbf{\Phi}_1 \odot \mathbf{\Psi}_2, ... , \mathbf{\Phi}_M \odot \mathbf{\Psi}_M) 
\label{eq:boxdotdictionary}
\end{equation}

One can now ask under what conditions Hadamard product and tensor product become mathematically equivalent, that is, can any sparse tensor product in (\ref{bind}) be uniquely inferred from the Hadamard product using a CS inference procedure (\ref{cs_recovery}). 
%With this interpretation of (\ref{bind}), the Hadamard product and tensor product binding become mathematically equivalent under certain conditions.
The following two lemmas consider a worst-case scenario in which there is equivalence between the atomic sparse and dense vectors, which requires that the sparks of the individual codebooks are at least $2K+1$. 
%This avoids the situation where the difference of any two sparse vectors can lie in the kernel of the codebooks, which would make the recovery impossible.

\vspace{0.5cm}
\noindent
{\it Lemma 1:} Let $\mbox{Spark}(\mathbf{\Phi}) = \mbox{Spark}(\mathbf{\Psi}) = 2K+1$. Then the spark of the sampling matrix in (\ref{bind}) is $\mbox{Spark}((\mathbf{\Phi} \boxdot \mathbf{\Psi})) \leq 2K+1$. 

\vspace{0.5cm}
\noindent
{\it Proof:} Choose a $(2K+1)$-sparse vector $\mathbf{c}$ in the kernel of $\mathbf{\Phi}$, and choose any cardinal vector $\mathbf{b}^j:=(0,...,0,1,0,...,0)$ with the nonzero component at index $j$. Then we have: $0 = \mathbf{\Phi} \mathbf{\alpha} = \mathbf{\Phi} \mathbf{c} \odot  \mathbf{\Psi}_{j} =  \sum_{i \in \mathbf{\alpha}} c_i \mathbf{\Phi}_{i} \odot  \mathbf{\Psi}_{j} = (\mathbf{\Phi} \boxdot \mathbf{\Psi}) \ vec( {\bf c} \otimes {\bf b}^j)$. Thus the $(2K+1)$-sparse vector $vec( {\bf c} \otimes {\bf b}^j)$ lies in the kernel of the sampling matrix in (\ref{bind}). There is also a small probability that the construction of $(\mathbf{\Phi} \boxdot \mathbf{\Psi})$ produces a set of columns with less than $2K+1$ components that are linearly dependent. 

\noindent
$\square$
\vspace{0.5cm}

Lemma 1 reveals that the sampling matrix (\ref{eq:boxdotdictionary}) does certainly not allow the recovery of $K^2$-sparse patterns in general. However, this is not required since
%the equivalence of Hadamard product and tensor product only requires that the outer products of $K$-sparse vectors can be inferred. T
the reshaped outer products of $K$-sparse vectors form a subset of $K^2$-sparse patterns. The following lemma shows that for this subset recovery can still be possible.  

\vspace{0.5cm}
\noindent
{\it Lemma 2:} The difference between the outer-products of pairs of $K$-sparse vectors cannot fully coincide in support with the $(2K+1)$-sparse vectors in the kernel of the sampling matrix of (\ref{bind}) as identified by Lemma 1. Thus, although $\mbox{Spark}((\mathbf{\Phi} \boxdot \mathbf{\Psi})) \leq 2K+1$, the recovery of reshaped tensor products from the Hadamard product can still be possible.

\vspace{0.5cm}
\noindent
{\it Proof:} The $(2K+1)$-sparse vectors in the kernel of the sampling matrix $(\mathbf{\Phi} \boxdot \mathbf{\Psi})$ identified in Lemma 1 
correspond to an outer product of a $(2K+1)$-sparse vector with a $1$-sparse vector. The resulting matrix has $2K+1$ nonzero components in one single column.  

The difference of two outer products of $K$-sparse vectors yields a matrix which can have maximally $2K$ nonzero components in one column. Thus, the sampling matrix should enable the unique inference of the tensor product from the Hadamard product of the dense vectors. 

\noindent
$\square$
\vspace{0.5cm}

\begin{figure}[h!]
    \centering
    \includegraphics[width=0.45\textwidth]{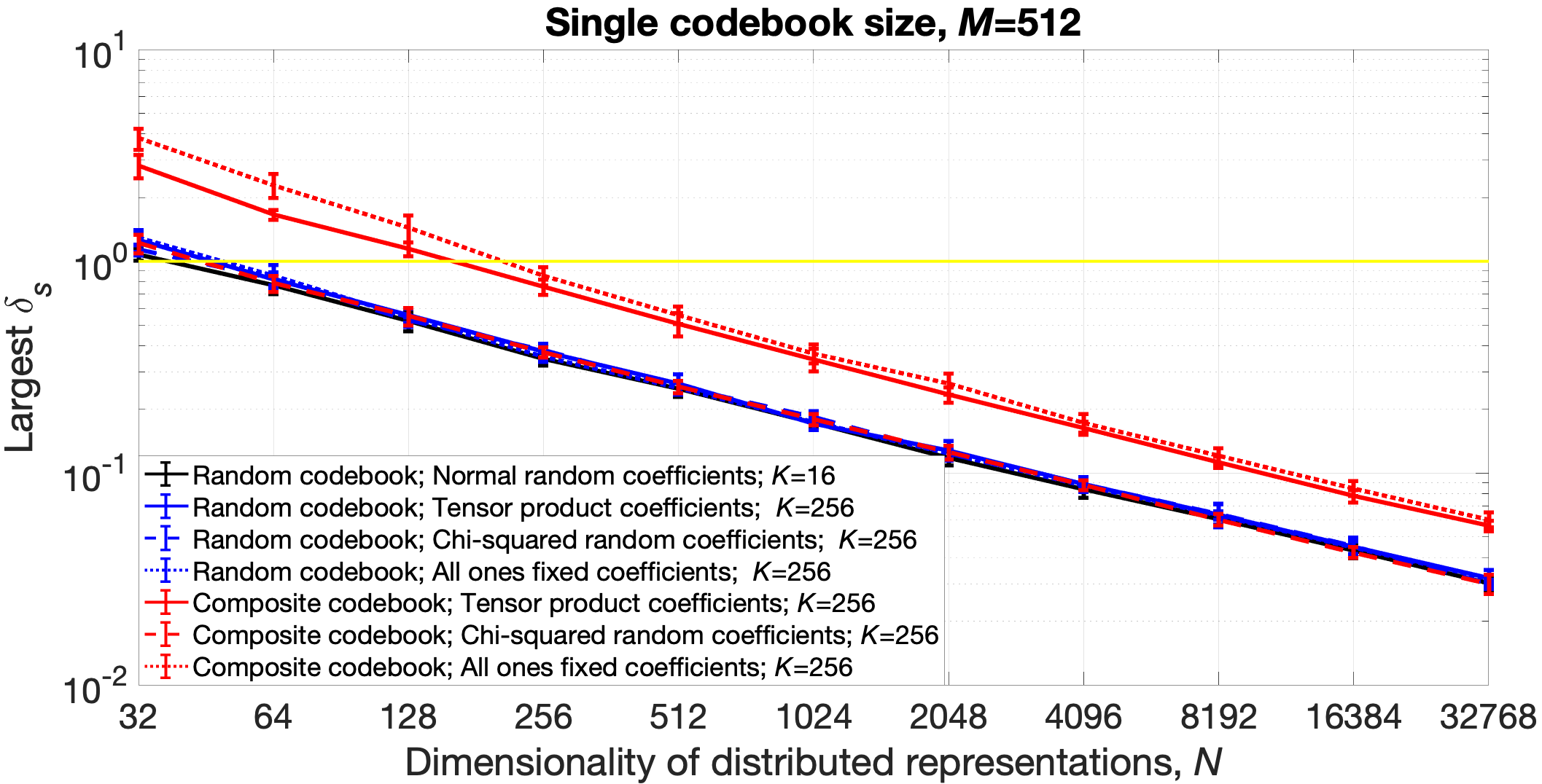}
    \caption{{\bf Worst-case RIP constant for inferring sparse tensor products in ensemble of random codebooks.} The largest empirical RIP constant ($\delta_s$) in an ensemble of 10 pairs of pseudo-random dictionaries $\mathbf{\Phi}, \mathbf{\Psi}$. For each pair the maximum RIP was determined by compressing 10000 sparse vectors. 
    For successful inference of the sparse representations, the RIP constant has to be below the $\delta_s=1$ level (yellow line). The black solid line represents the RIP for inferring atomic sparse vectors from dense vectors formed according to (\ref{cs_compression}).
    The red solid line represents the RIP for inferring tensor products from dense vectors formed according to (\ref{bind}).
    Other lines in the diagram are controls. 
    The blue solid line represents the RIP for a ($N\times M^2$) random dictionary in which all elements are indpendently sampled rather than constructed by $\mathbf{\Phi} \boxdot \mathbf{\Psi}$ from the smaller dictionaries. Dashed and dotted red lines represent the RIP using the $\mathbf{\Phi} \boxdot \mathbf{\Psi}$ sampling matrix with sparse vectors with independent random components, rather than formed by a tensor product $\ vec( {\bf a} \; {\bf b}^{\top})$ of two random vectors. The blue dashed line is for real-valued vectors with elements sampled from a from a Chi-squared distribution, the blue dotted line for binary random vectors. Dashed and dotted blue lines represent the RIP for the same type of independent random vectors with the independent random sampling matrix. 
    %The controls for Chi-squared distribution (red dashed line) and binary (red dotted line) sparse vectors are also depicted for $\mathbf{\Phi} \boxdot \mathbf{\Psi}$.
    %The blue solid line dispays a control, where the tensor product was sampled with a random $N\times M^2$ , rather than with the dictionary as in (\ref{eq:boxdotdictionary}). 
    %Dotted lines show the RIP values for inferring representations involving binary rather than continuous-valued sparse vectors, atomic vectors (blue) and tensor products (red). 
    %The solid blue line shows the RIP condition for inferring the sparse tensor product from a dense representation, sampled with a fully random 
    %For comparison, the solid black line depicts the RIP constant for the atomic vectors from $N\times M$ dictionary.
    %The dashed lines 
    }
    \label{fig:res_binding}
\end{figure}

Lemmas 1 and 2 investigate the equivalence of Hadamard and tensor product binding in the worst case, that is, when the
codebooks have the minimum spark that still guarantees the unique equivalence between the sparse and dense atomic vectors. To explore the equivalence in the case of random codebooks, we performed simulation experiments with a large ensemble of randomly generated codebook pairs $(\mathbf{\Phi}, \mathbf{\Psi})$. Fig.~\ref{fig:res_binding} shows the averaged worst (i.e., highest) RIP constant amongst the ensembles for inferring the tensor product from the Hadamard product (solid red line). 

Compared to the RIP constant for inferring the sparse representations of atomic vectors (black line), the RIP constant for inferring the tensor product (red line) is significantly higher. Thus, tensor product and Hadamard product are not always equivalent even if the atomic sparse and dense vectors are equivalent -- in the example, when the dimension of dense vectors is between $N=40$ to $N=140$. However, with the dimension of dense vectors large enough ($N>140$), the equivalence holds. Further, the controls in Fig.~\ref{fig:res_binding} help to explain the reasons for the gap in equivalence for small dense vectors. The RIP constants are significantly reduced if the tensor product is subsampled with a fully randomized matrix (solid blue line), rather than with the sampling matrix resulting from (\ref{bind}). In contrast, the requirement to infer outer products of continuous valued random vectors (solid red line) does not much increase the RIP values over the RIP requirement for the inference of outer products of binary vectors (dotted red line). Thus, we conclude that sampling with matrix $\mathbf{\Phi} \boxdot \mathbf{\Psi}$ (\ref{eq:boxdotdictionary}), which is not i.i.d. random but formed by a deterministic function from the smaller atomic random sampling matrices, requires a somewhat bigger dimension of the dense vectors to be invertible.

Here we have shown that under certain circumstances the binding operation between dense vectors in the MAP VSA is mathematically equivalent to the tensor product between the corresponding sparse vectors. 
This equivalence reveals a natural link between two prominent proposals for symbolic binding in the literature, the dimensionality preserving binding operations in VSA models, with the tensor product in the TPR model \citep{Smolensky1990, smolensky2016}. 
In other VSA models, such as HRR \citep{Plate2003}, atomic symbols are represented by dense Gaussian vectors and the binding operation is circular convolution. Our treatment can be extended to these models by simply noting that by the Fourier convolution theorem (\ref{fct}) circular convolution is equivalent to the Hadamard product in the Fourier domain, i.e. $\mathbf{x} \ast  \mathbf{y} = \mathcal{F}^{-1}\left(\mathcal{F}(\mathbf{x}) \odot \mathcal{F}(\mathbf{y})\right)$.

\subsubsection{Set operations}
\label{sec:protsum}
%\label{sec:simple_addition}

Summing dense vectors corresponds to summing the sparse vectors
\begin{equation}
{\bf x}+{\bf y} = \mathbf{\Phi} ({\bf a}+{\bf b})
\label{eqn:addition}
\end{equation}
Thus, the sum operation represents a bag of features from all objects, but the grouping information on how these features were configured in the individual objects is lost. The inability to recover the individual compound objects from the sum representation has been referred to as the binding problem in neuroscience (\cite{treisman1998}).

The protected sum of set vectors (\ref{prot_sum}) can resolve the binding problem. This relies on binding the dense representations of the individual objects to a set of random vectors that act as keys, stored in the codebook $\mathbf{\Psi}$ (\ref{prot_sum}):
%It can be computed from the matrix of all sparse feature vectors of the individual sets: 
%If the objects are again composed of sparse sets of features, 
%encodes corresponding sparse coefficient vectors $\{ {\mathbf{a}^j}, j=1,..., L \}$ using the codebook $\mathbf{\Phi}$: $\mathbf{x}^j = \mathbf{\Phi} \mathbf{a}^j$. the protected sum becomes:
%The codebook $\mathbf{\Phi}$ and the sparse coefficient vectors satisfy the CS condition (\ref{cs_cond}). 
%Using another random codebook $\mathbf{\Psi}$, independently chosen from $\mathbf{\Phi}$, the protected sum of the dense vectors can be formed as 
\begin{equation}
%\begin{align}
%\begin{split}
%\mathbf{s}
\sum_j^L \mathbf{\Psi}_j \odot \mathbf{x}^j = \sum_j^L \mathbf{\Psi}_j \odot \sum_i^M \mathbf{\Phi}_i a_i^j
    = (\mathbf{\Phi} \boxdot \mathbf{\Psi}) (\mathbf{a}^1, \mathbf{a}^2, ..., \mathbf{a}^L)
%    &= (\mathbf{\Phi} \boxdot \mathbf{\Psi}) \  vec( \mathbf{A})
%{\bf x}+{\bf y} = \mathbf{\Phi} {\bf a} + \mathbf{\Psi} {\bf b} = [\mathbf{\Phi}, \mathbf{\Psi}](\mathbf{a}, \mathbf{b})
\label{eqn:set}
%\end{split}
%\end{align}
\end{equation}
This shows that the protected sum can be computed from the concatenation of sparse vectors. The concatenation of sparse vectors is a representation that fully contains the binding information, but again leads to an increase in dimensionality. Similar to (\ref{bind}), (\ref{eqn:set}) describes linear sampling of a sparse vector like in  compressed sensing. The sampling matrix $(\mathbf{\Phi} \boxdot \mathbf{\Psi})$ is a $N \times ML$ sampling matrix formed by each pair of vectors in $\mathbf{\Phi}$ and $\mathbf{\Psi}$, as in (\ref{eq:boxdotdictionary}), and the sparse vector is the $ML$-dimensional concatenation vector.
%$(\mathbf{a}^1, \mathbf{a}^2, ..., \mathbf{a}^L)$ is the concatenation of the individual sparse vectors $\mathbf{a}^j$. The concatenated coefficient vector may also be written as matrix $\mathbf{A}$, whose columns contain the coefficient vectors.

We again ask under what conditions the sparse concatenation vector can be uniquely inferred given the dense representation of the protected sum, which makes the dense and sparse representations equivalent. Like in section~\ref{hadamard_tensor}, we first look at the worst case scenario, and then perform an experiment with codebooks composed of random vectors.
The worst case scenario assumes the spark of $\mathbf{\Phi}$ to be $2K+1$, just big enough that atomic vectors can be inferred uniquely. By Lemma 1, the spark of the sampling matrix is smaller or equal to $2K+1$, smaller than the sparsity $KL$ of vectors to be inferred. Again, the vectors to be inferred are a subset of $KL$-sparse vectors, the vectors that have $K$ nonzero components in each of the $L$ $M$-sized compartments. Thus, as in Lemma 2 for the Hadamard product, the difference formed by two of these vectors can maximally produce $2K$ nonzero components in each compartment, and therefore never coincide with a kernel vector of the sampling matrix.

%The map from sparse to dense vectors given by (\ref{eqn:set}) again has the form of a matrix-vector multiplication where the vector is sparse and the matrix is randomized. If the CS condition (\ref{cs_cond}) is fulfilled, then there is bijective unique map between the protected sum of dense vectors and the concatenation of corresponding sparse vectors. This means in inequality (\ref{cs_cond}), $M$ is replaced by $ML$, and $K$ by $KL$. Otherwise, the sum of dense vectors represents a lossy compressed version of the concatenation of the corresponding sparse vectors.

\begin{figure}[tb]%[h]
    \centering
    \includegraphics[width=0.45\textwidth]{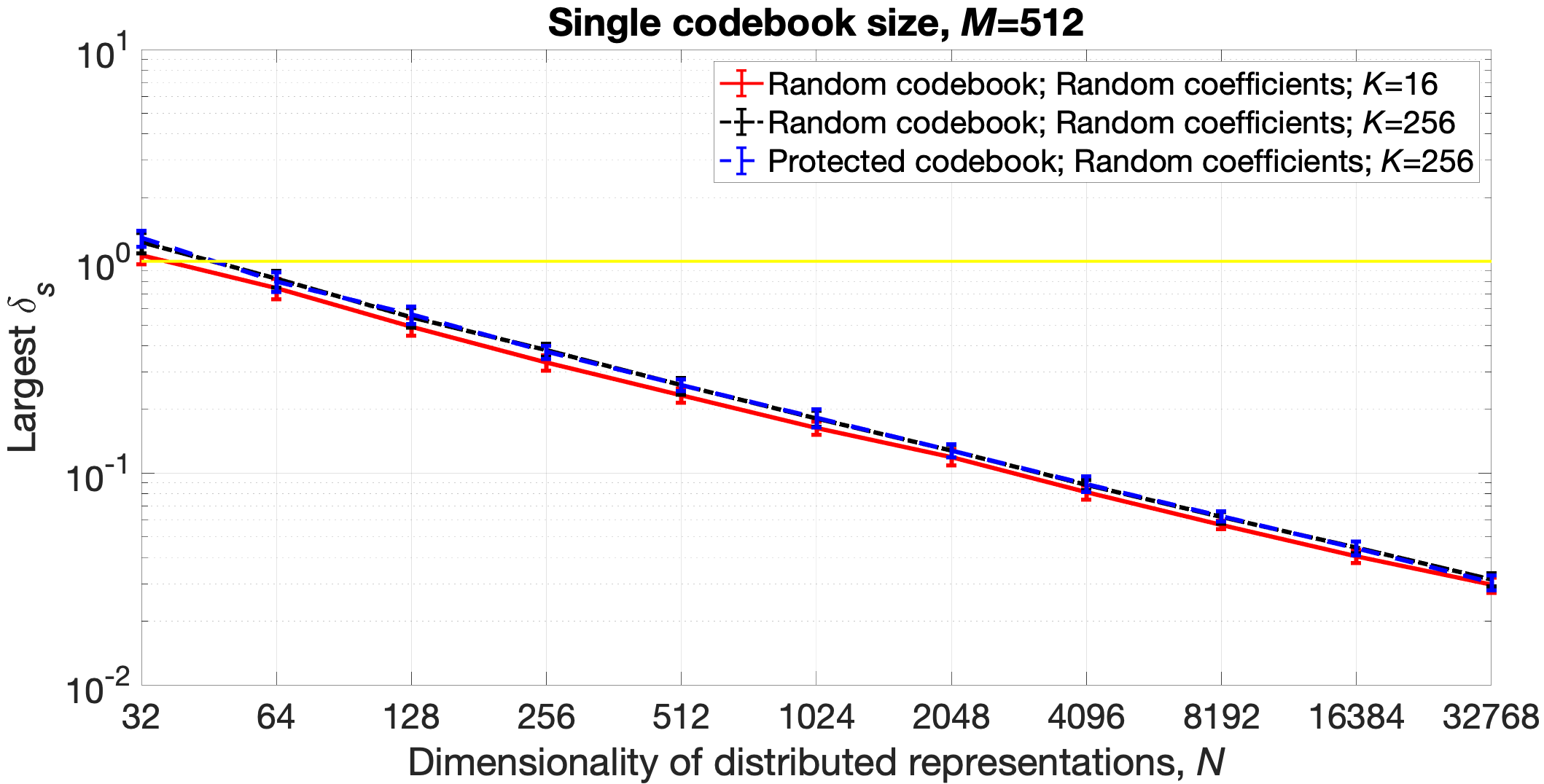}
    \caption{{\bf Worst-case RIP constant for inferring sparse representations of protected sums in ensemble of random codebooks.} The largest empirical RIP constant ($\delta_s$) in ensemble of 10 pairs of pseudo-random dictionaries $\mathbf{\Phi}, \mathbf{\Psi}$. For each pair the maximum RIP was determined by compressing 10000 sparse vectors. 
    For successful inference of the sparse representations, the RIP constant has to be below the $\delta_s=1$ level (yellow line).
    Red solid line represents RIP values for inferring atomic sparse vectors from dense vectors formed according to (\ref{cs_compression}).
    Blue dashed line represents RIP values for inferring protected sum from dense vectors formed according to (\ref{eqn:set}).
    For comparison, black dotted line represents RIP values for inferring protected sum when instead of $\mathbf{\Phi} \boxdot \mathbf{\Psi}$ the dictionary is random. 
    %Note that when the RIP condition is fulfilled for the atomic vectors, the RIP is also almost fulfilled for the protected sum representations. Compared to variable binding, the gap between the RIP conditions of atomic vectors versus composite vectors for the protected sum is much smaller.
    }
    \label{fig:res_prot_sum}
\end{figure}
Fig.~\ref{fig:res_prot_sum} shows the results of simulation experiments with an ensemble of random codebooks. For the protected sum, the worst RIP values of the inference of individual sparse vectors versus the list of sparse vectors composing the protected sum do coincide. Thus, the dense protected sum vector and the list of sparse feature vectors are equivalent.  

The alternative method of forming a protected sum (\ref{eqn:permutprot_sum}) using powers of a permutation matrix $\mathbf{P}$, corresponds equally to a sampling of the concatenation of the sparse vectors. As long as the sampling matrices $(\mathbf{\Phi}, \mathbf{P \Phi}, \mathbf{P}^2\mathbf{\Phi}, ..., \mathbf{P}^{(L-1)}\mathbf{\Phi})$ and $(\mathbf{\Phi} \boxdot \mathbf{\Psi})$ have similar properties, the conditions for equivalence between protected sum and concatenated sparse vectors hold. 
%Generally, the permutation operation also acts as a randomizing operation, as long as the powers of the permutation are small compared to $N$.

\begin{figure*}[h]
\centering
%\hspace{2cm}
%\includegraphics[width=18cm]{spm_circuit.png}
%\vspace{-3.5cm}
%\includegraphics[width=0.4\textwidth]{binding_network-plate2000-191231.png}
\includegraphics[width=0.8\textwidth]{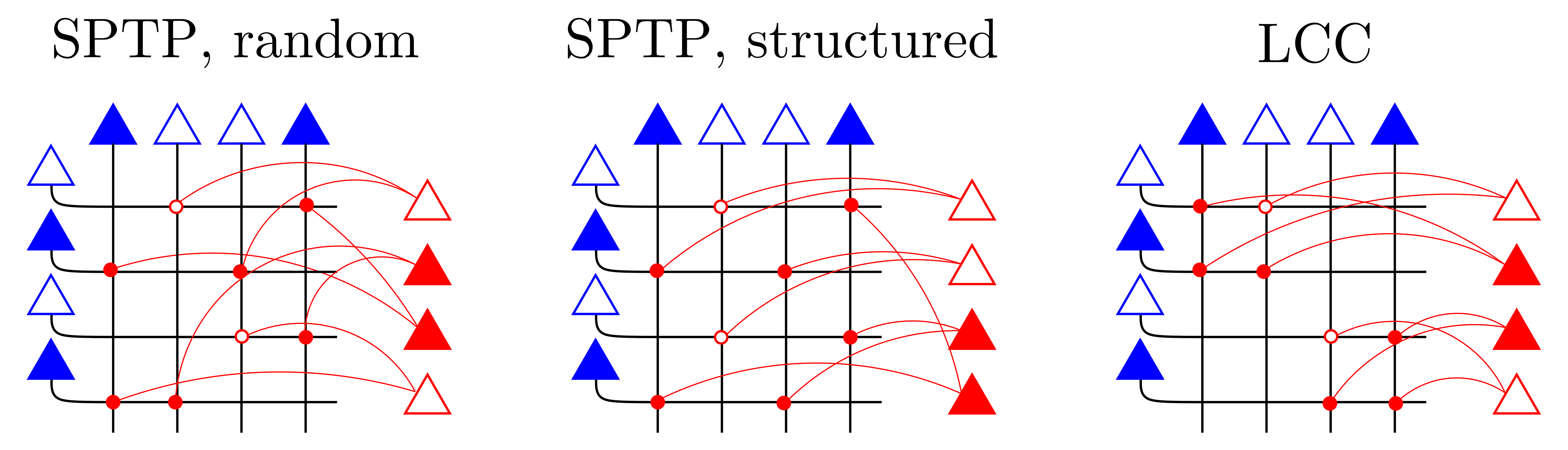}

\caption{ {\bf Circuits for sparsity-preserving binding}: Three pools of neurons (blue: two inputs, red: output) represent the sparse neural activity patterns $\mathbf{a}$, $\mathbf{b}$ and $\mathbf{c}$. The dendritic tree of the output neurons contains coincidence detectors that detect pairs of co-active axons (red circles), and the soma (red triangles) sums up several coincidence detectors based on the required fan-in. Each neuron samples only a subset of the outer product depending on the desired sparsity and threshold settings. The subsampling pattern of neurons is described by a binary tensor $W^l_{ij} \in \{0,1\}$, where $i,j$ indexes the coincidence point and $l$ the postsynaptic neuron. We examine three different sampling strategies random sampling, structured sampling, and the block-code.}
\label{model}
\end{figure*}

\subsection{Dimension- and sparsity-preserving VSA operations}
\label{sec:dso}
The results from Sect.~\ref{sec:cs} reveal that variable binding and the protected set representation in classical VSA models induce equivalent operations between sparse vectors that are not dimensionality preserving. Thus, dimensionality-preserving operations for binding and protected sum involve potentially lossy transformations of the higher dimensional data structure into a single vector. However, dimensionality-preserving binding operations have only been defined for dense VSA representations. In the following, we investigate binding operations on sparse VSA representations that are both dimensionality- and sparsity-preserving, one for general sparse vectors and one for sparse vectors with block structure.

\subsubsection{Sparsity-preserving binding for general $K$-sparse vectors}

%In section~\ref{hadamard_tensor}, we described two common methods of semantic binding in the literature, the Hadamard product and the tensor product. Both of these binding methods are ill-suited for recursive computations with sparse vectors. The Hadamard product between sparse vectors leads to sparser and sparser representations the more hierarchical the compound object becomes. Conversely, the tensor product produces a representation of the compound object whose dimension grows exponentially with each conjunction of atomic features. VSA methods for dimension-preserving binding can be considered as a compression step after forming the tensor product. The representation vector is formed by linear summation of a fixed deterministic sampling scheme of the tensor.

Binding operations in VSAs can all be described as a projection of the tensor product to a vector, including Hadamard product, circular convolution binding  \citep{Plate2003} and vector-derived transformation binding (VDTB) \citep{gosmann2019}, see Appendix (\ref{tensor_product_projection}).  
 %For instance, each output neuron in circular convolution sums a diagonal of the tensor product. 
However, when applied to sparse atomic vectors, these operations do not preserve sparsity -- circular convolution produces a vector with reduced sparsity, while the Hadamard product increases sparsity. 
%In the next section, we investigate strategies to compress the tensor product while maintaining sparsity. 

Ideally, a sparsity-preserving VSA binding operation operates on two atomic vectors that are $K$-sparse and produces a $K$-sparse vector that has the correct algebraic properties. To preserve sparsity, we developed a binding operation that is a projection from a sub-sampling of the tensor product. We refer to this operation as {\it sparsity-preserving tensor projection (SPTP)}.
Given two $K$-sparse binary vectors $\mathbf{a}$ and $\mathbf{b}$, SPTP variable binding 
%$\mathbf{c} = \mathbf{a} \  \underline{\times} \  \mathbf{b}$
%$\mathbf{c} = \mathbf{a} \odot \mathbf{b}$ [DO WE NEED A NEW OPERATION SYMBOL?]
is given by:
\begin{equation}
    %\mathbf{a} \circ \mathbf{b} := 
    (\mathbf{a} \  \underline{\circ} \  \mathbf{b})_l = H\left(\sum_{ij} W^l_{ij} a_i b_j -\theta \right)
    % for phasor version:
    % z = \sum_{ij} W^l_{ij} a_i b_j
    % \mathbf{a} \underline{\circ} \mathbf{b} = z/|z| H(|z| - \theta)
    \label{sptp}
\end{equation}
Here $H(x)$ is the Heaviside function, $\theta$ is a threshold. 
For a pair of $K$-sparse complex phasor vectors, SPTP binding is defined as:
\begin{eqnarray}
    %\mathbf{a} \circ \mathbf{b} := \mathbf{a} \  %\underline{\circ} \  \mathbf{b} = H\left(\sum_{ij} W^l_{ij} %a_i b_j -\theta \right)
    % for phasor version:
     (\mathbf{a} \ \underline{\circ} \ \mathbf{b})_l &=& \frac{z_l}{|z_l|} H(|z_l| - \theta)
    \label{sptp_c}\\
    z_l &=& \sum_{ij} W^l_{ij} a_i b_j\nonumber
\end{eqnarray}

The computation of (\ref{sptp}) resembles a circuit of threshold neurons with coincidence detectors in their dendritic trees, see Fig.~\ref{model}. 
The synaptic tensor $\mathbf{W} \in \{0,1\}^{M \times M \times M}$ is a binary third-order tensor that indicates how each output neuron samples from the outer-product. We examined two types of sampling tensors, one with the $1$-entries chosen i.i.d. (without repetition), and one with $1$-entries aligned along truncated diagonals of the tensor (left and middle panel in Fig.~\ref{model}).
%, see Appendix~\ref{sec:classical_vsa_binding}. 

The sparsity of the output in (\ref{sptp}) is controlled by the threshold and by the density of this sampling tensor.  
To achieve a target sparsity of $K/N$ for a threshold $\theta=1$, the fan-in to each neuron has to be $N/K$ (see analysis in Appendix~\ref{sec:fan-in}).   
%is set based on the threshold such that the output vector will be $K$-sparse, which can be computed from binomial statistics . For 
%, adjusted so that the sparsity of the $\mathbf{a} \  \underline{\circ} \  \mathbf{b}$ is also close to $K$. 
%, connecting a downstream neuron $l$ to a two-dimensional index field $i \times j$. Each pair $i,j$ indexes one component of the outer product matrix between $\mathbf{a} \otimes \mathbf{b}$.  
%In the following we summarize the results of the analysis.  
%We describe the required sampling density of the tensor $W_{ij}^l$ based on different threshold settings, and how to unbind in the Appendix.
Thus, the minimal fan-in of the sampling tensor $\mathbf{W}$ increases with sparsity. If the pattern activity is linear in the dimension, $K=\beta N$ with $\beta << 1$, the minimal fan-in is $\alpha^* = 1/\beta$. In this case, the computational cost of SPTP binding is order $N$. If the pattern activity goes with the square root of the dimension, $K=\beta \sqrt{N}$, the minimal fan-in is $\alpha^* = \sqrt{N}/\beta$. If the pattern activity goes with the logarithm of the dimension,  $K=\beta \ln(N)$, the minimal fan-in is $\alpha^* = N/(\beta\ln(N))$.  
%Thus, the sparser the patterns, the bigger the minimal fan-in.
Further, for optimizing the unbinding performance, the sampling tensor should fulfill the symmetry condition $W^i_{jl} = W^l_{ij}$ (see analysis in Appendix~\ref{sec:appendix_symmetry}).
%According to this analysis, the fan-in of CCB should be equally able to support the sparsities described above, VDTB should be able to support linear and square root sparsity. 

\begin{figure*}[h]
    \centering
    \includegraphics[width=0.9\textwidth]{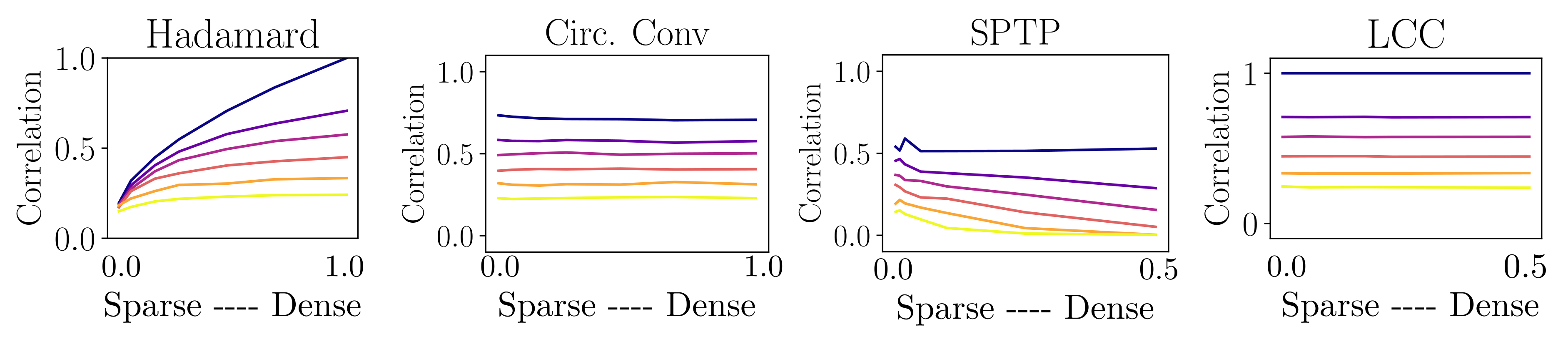}
    \caption{\textbf{Comparison of binding operations.} The unbinding performance was measured as the correlation between ground truth and output of unbinding. Different levels of sparsity (x-axis) and superposition were examined (colored lines: [0, 1, 2, 4, 8 16] items in superposition).}
    \label{fig:comparison}
\end{figure*}

\subsubsection{Sparsity-preserving binding for sparse block-codes}

We next consider sparse vector representations that are constrained as block-codes \citep{gripon2011}, which have been proposed for VSAs before \citep{Laiho2015}. 
Our model extends this previous work with a block-code in the complex domain.
In a sparse block-code, a vector of length $N$ is divided into $K$ equally-sized blocks, each with a one-hot component. In the complex domain, the hot component is a phasor with unit amplitude and arbitrary phase.%or a single neuron that ``spikes''. 

%The addition operation is  a regular vector addition. As we saw before, such a set operation preserves dimension but not sparsity. In the case of sparse block vectors, the resulting vector of a sum contains more than one active component in the blocks. 

The binding operation \citet{Laiho2015} proposed operates on each block individually.
For each block, the indices of the two active elements of the input are summed modulo block size to produce the index for the active element of the output. 
%is the addition of the index numbers of the active elements in each block modulo block size,
This is the same as circular convolution \citep{Plate1994} performed locally between individual blocks. This binding operation, \emph{local circular convolution} (LCC), denoted by $\ast_b$, produces a sparse block-code when the input vectors are sparse block-codes (Fig.~\ref{model}). LCC variable binding can be implemented by forming the outer product and sampling as in Fig. \ref{model}, with a circuitry in which each neuron has a fan-in of $\alpha=N/K$ and samples along truncated diagonals of the tensor product. LCC has a computational complexity of $\alpha N$, which is order $N$ if $K$ is proportional to $N$. An alternative implementation (that is more efficient on a CPU) uses the Fourier convolution theorem (\ref{fct}) to replace convolution by the Hadamard product:
\begin{align}
    \begin{split}
    (\mathbf{a} \ast_b \mathbf{b})_{block_i} &= 
    \mathbf{a}_{block_i} \ast \mathbf{b}_{block_i} \\
    &= \mathcal{F}^{-1}\left(\mathcal{F}(\mathbf{a}_{block_i}) \odot \mathcal{F}(\mathbf{b}_{block_i})\right)
\end{split}
\end{align}
where $\mathcal{F}$ is the Fourier transform.
%The block-code is guaranteed to have only one active element in each block, which also guarantees that the circular convolution operation on each block  also produces a one-hot vector . 

The LCC unbinding of a block can be performed by computing the inverse of the input vector to unbind. This is the inverse with respect to circular convolution, which is computed for each block, 
\begin{equation}
    \mathbf{a}_{block_i}^{-1}= \mathcal{F}^{-1}(\mathcal{F}(\mathbf{a}_{block_i})^{*})
    \label{eqn:lcc_inv}
\end{equation}
where $^{*}$ is the complex conjugate. The inverse is used when unbinding, for instance, if $\mathbf{c} = \mathbf{a} \ast_b \mathbf{b}$, then $\mathbf{a} = \mathbf{b}^{-1} \ast_b \mathbf{c}$. 
%$\mathbf{c} = \mathbf{a} \boxast \mathbf{b}$, then $\mathbf{a} = \mathbf{b}^{-1} \boxast \mathbf{c}$. 

\subsubsection{Experiments with sparsity-preserving binding}

The binding operations are evaluated based on whether they maintain sparsity and how much information is retained when unbinding.
Circular convolution and Hadamard product can be ruled out, because they do not preserve sparsity, the Hadamard product increases sparsity, and circular convolution reduces sparsity, but we still evaluated these operations for comparison.

\begin{figure}[tb]%[h]
    \centering
    \includegraphics[width=0.45\textwidth]{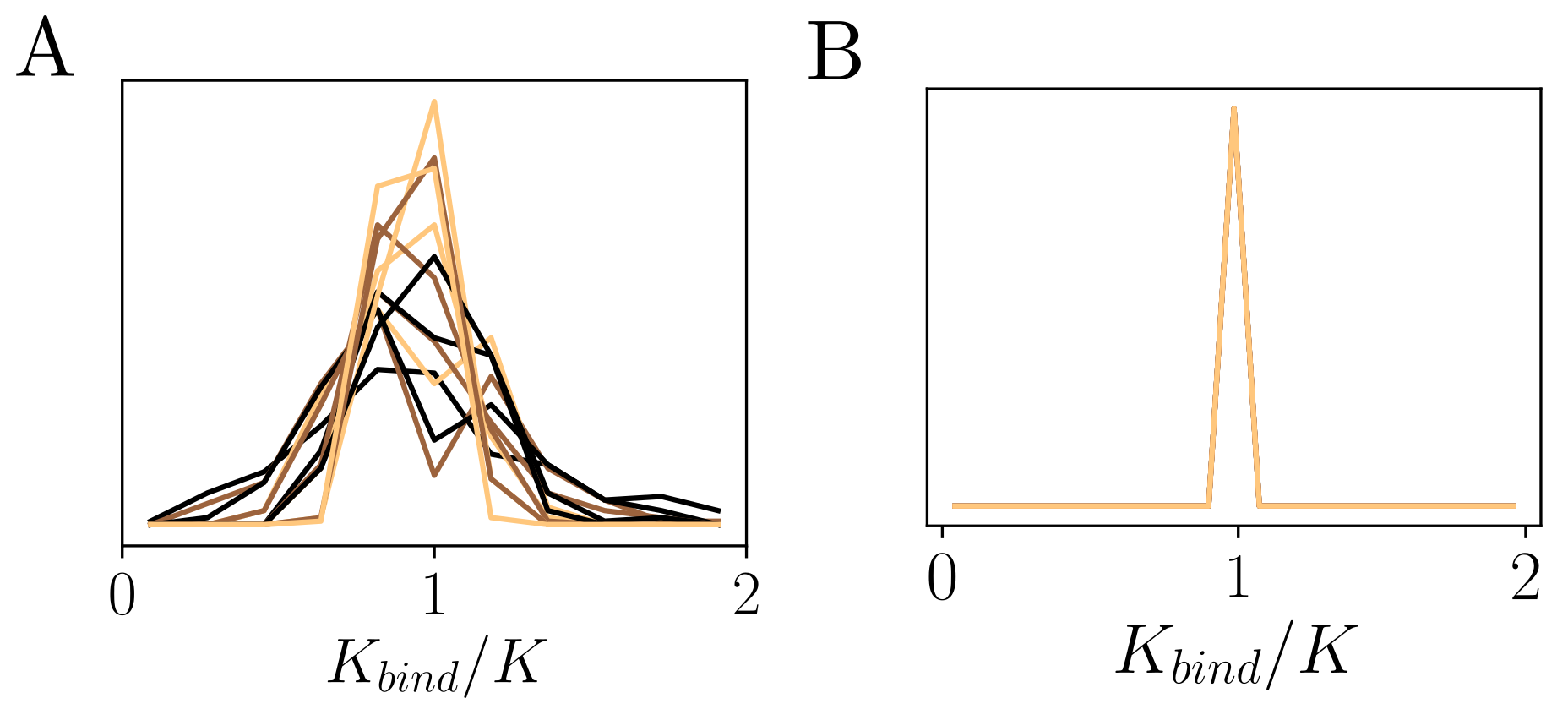}
    \caption{\textbf{Preservation of sparsity with a binding operation.} A. The output sparsity $K_{bind}$ is compared to the sparsity of the base vectors $K$. Binding with SPTP results in an output vector that has the correct expected sparsity, but there is some random variance. This variance reduces with more active components ($K=[20, 50, 100, 200]$ black to orange lines). This result is similar for both random and structured SPTP. B. The output sparsity of binding sparse block-codes with LCC deterministically results in a vector which maintains the sparsity of the inputs.}
    \label{fig:maintain_sparse}
\end{figure}

We investigated how well sparsity is preserved with LCC and SPTP binding (Fig.~\ref{fig:maintain_sparse}). We find that LCC binding preserves sparsity perfectly, and SPTP binding preserves sparsity on average (statistically), but with some variance.
%The binding operations we propose for sparse vectors both maintain sparsity of the inputs.
%However, both SPTP strategies result in only preserving sparsity statistically (Fig.~\ref{fig:maintain_sparse}A). 

We next measure how much information is retained when first binding and then unbinding a vector using the proposed binding operations (Fig.~\ref{fig:comparison}). The Hadamard product binding achieves the highest correlation values for dense vectors, but performs very poorly for sparse vectors. The other three binding methods perform equally across sparsity levels. Circular convolution and SPTP binding are somewhat lossy for all sparsity levels. The LCC variable binding between block-codes achieves the highest correlation values, outperforming circular convolution, and SPTP binding.  Each diagram in Fig.~\ref{fig:comparison} contains 6 curves, corresponding to different levels of additive superposition in the bound vectors.

%We also analyze as a baseline comparison Hadamard binding and circular convolution. 

SPTP binding works for general $K$-sparse vectors. It has decent properties but is somewhat lossy. The information loss is due to the fact that not all active input components contribute to the generation of active outputs, which means that some active input components cannot be inferred during unbinding and information is lost. The loss can be kept at a minimum by using a synaptic weight tensor that fulfills the symmetry condition $W^i_{jl} = W^l_{ij}$. This information loss persisted regardless of SPTP being structured or random, or the threshold and fan-in settings.

These experiments identify LCC binding as an ideal sparsity-preserving binding operation. With sparse block-codes and local circular convolution applied separately to each block, the unbinding is loss-less. The block structure guarantees that each active input component participates in the formation of an active output component, which cannot be guaranteed for general $K$-sparse vectors.  
Of course, there is a price to pay, LCC binding requires the atomic vectors to be sparse block-codes. The coding entropy of block-codes is significantly smaller than general $K$-sparse patterns.

%This results in an ideal binding operation that fulfills all requirements and maximizes information retained when unbinding. 

\subsection{Applications of VSAs with sparse block-codes}
\label{sec:apps}
\subsubsection{Solving symbolic reasoning problems}

As a basic illustration of symbolic reasoning with sparse block-codes, we implement the solution to the cognitive reasoning problem \citep{Kanerva2010}: ``What's the dollar of Mexico?'' in the supplemental Jupyter notebook. To answer such queries, data structures are encoded into vectors that represent trivia information about different countries. A data record of a country is a table of key-value pairs. For example, to answer the specific query, the relevant records are:
%\emph{USA} and \emph{Mexico} have to be selected:  
\begin{eqnarray}
\mathbf{ustates} &=& \mathbf{nam} \ast_b \mathbf{usa} + \mathbf{cap} \ast_b \mathbf{wdc} + \mathbf{cur} \ast_b \mathbf{dol}\nonumber\\
\mathbf{mexico} &=& \mathbf{nam} \ast_b \mathbf{mex} + \mathbf{cap} \ast_b \mathbf{mxc} + \mathbf{cur} \ast_b \mathbf{pes}\nonumber
\end{eqnarray}
The keys of the fields \emph{country name}, \emph{capital} and \emph{currency} are represented by random sparse block-code vectors $\mathbf{nam}$, $\mathbf{cap}$ and $\mathbf{cur}$.
%, respectively, and act as keys. 
The corresponding values \emph{USA}, \emph{Washington D.C.}, \emph{Dollar}, \emph{Mexico}, \emph{Mexico City}, and \emph{Peso} are also represented by sparse block-code vectors $\mathbf{usa}$, $\mathbf{wdc}$, $\mathbf{dol}$, $\mathbf{mex}$, $\mathbf{mxc}$, $\mathbf{pes}$.
%respectively, and act as values. 
All the vectors are stored in the codebook $\mathbf{\Phi}$. 
The vectors $\mathbf{ustates}$ and $\mathbf{mexico}$ represent the complete data records --  they are a representation of key-value pairs that can be manipulated to answer queries. These record vectors have several terms added together, which reduces the sparsity.

To perform the reasoning operations required to answer the query, first the two relevant records have to be retrieved in the database. While $\mathbf{mexico}$ can be found by simple pattern matching between terms in the query and stored data record vectors, the retrieval of $\mathbf{ustates}$ is not trivial. 
%It requires pattern matching of the terms in the query with the set of data records unbound by key vectors.
%A possible solution to perform this search in parallel are resonator networks \citep{kent2019resonator}. 
The original work does not deal with the language challenge of inferring that the $\mathbf{ustates}$ record is needed. Rather, the problem is formally expressed as analogical reasoning, where the query is given as: \emph{Dollar}:\emph{USA}::?:\emph{Mexico}. Thus, the pair of records needed for reasoning are given by the query.

Once the pair of records is identified, the following transformation vector is created:
$$
\mathbf{t}_{UM} = \mathbf{mexico} \ast_b \mathbf{ustates}^{-1}
$$
Note that unbinding with LCC is to bind with the inverse vector (\ref{eqn:lcc_inv}), whereas  
%binding operation is needed to form $\mathbf{t}_{UM}$.
in the MAP VSA used in the original work  \citep{Kanerva2010} the binding and unbinding operations are the same. 
%vectors are their own self-inverses). 
The transformation vector will also contain many summed terms, leading to less sparsity.
The transformation vector then contains the relationships between the different concepts
$$
\mathbf{t}_{UM} = \mathbf{mex} \ast_b \mathbf{usa}^{-1} + \mathbf{mxc} \ast_b \mathbf{wdc} ^ {-1} + \mathbf{pes} \ast_b \mathbf{dol}^{-1} + noise
$$
where all of the cross-terms can be ignored and act as small amounts of cross-talk noise. 

The correspondence to dollar can be computed by binding $\mathbf{dol}$ to the transformation vector:
%\begin{align*}
 %   \begin{split}
$$\mathbf{ans} = \mathbf{dol} \ast_b \mathbf{t}_{UM} = \mathbf{pes} + noise$$
  %  \end{split}
%\end{align*}

The vector $\mathbf{ans}$ is then compared to each vector in the codebook $\mathbf{\Phi}$. The codebook entry with highest similarity represents the answer to the query. This will be \emph{Peso} with high probability for large $N$.
The probability of the correct answer can be understood through the capacity theory of distributed representations described in \citet{Frady2018}, which we next apply to this context.

In general, a vector like $\mathbf{t}_{UM}$ can be considered as a mapping between the fields in the two tables. The number of entries will determine the amount of crosstalk noise, but all of the entries that are non-sensible also are considered crosstalk noise.

Specifically, we consider general data records of key-value pairs, similar in form to $\mathbf{ustates}$ and $\mathbf{mexico}$. These data records will contain $R$ ``role'' vectors that act as keys.  Each one has corresponding $M_r$ potential ``filler'' values. The role vectors are stored in a codebook $\mathbf{\Psi} \in \mathbb{C}^{N \times R}$. For simplicity, we assume that all $R$ roles are present in a data record, each with one of the $M_r$ fillers attached. The fillers for each role are stored in the codebook 
${\mathbf{\Phi}}^{(r)} \in \mathbb{C}^{N \times M_r}$. This yields a generic key-value data record:
\begin{equation}
\mathbf{rec} = \sum_r^R \mathbf{\Psi}_r \ast_b {\mathbf{\Phi}}^{(r)}_{i^*}
\end{equation}
where the index $i^*$ indicates one filler vector from the codebook for a particular role.

Next, we form the transformation vector, which is used to map one data record to another. This is done generically by binding two record vectors: $\mathbf{t}_{ij} = \mathbf{rec}_j \ast_b \mathbf{rec}_i^{-1}$.

As discussed, the terms in each record will distribute, and the values that share the same roles will be associated with each other. But, there are many cross-terms that are also present in the transformation vector that are not useful for any analogical reasoning query. The crosstalk noise is dependent on how many terms are present in the sum, and this includes the cross-terms. Thus, the total number of terms in the transformation vector $\mathbf{t}_{ij}$ will be $R^2$.

In the next step, a particular filler is queried and the result is decoded by comparison to the codebook $\mathbf{\Phi}$, which contains the sparse-block code of each possible filler: 
\begin{equation}
\mathbf{a}_r = \mathbf{\Phi}^{(r)} (\mathbf{t}_{ij} \ast_b \mathbf{\Phi}^{(r)}_{j^*})
\end{equation}
where $j^*$ indicates the index of the filler in the query (e.g. the index of \emph{Dollar}).
The entry with the largest amplitude in the vector $\mathbf{a}_r$ is considered the output.

\begin{figure}
    \centering
    \includegraphics[width=0.3\textwidth]{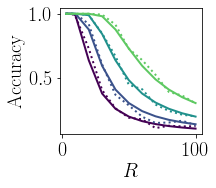}
    \caption{\textbf{Performance of analogic reasoning tasks with sparse block-codes.} We empirically simulated analogic reasoning tasks with data records containing $R$ key-value pairs, and measured the performance (dashed lines). This performance can be predicted based on the VSA capacity theory reported in \citet{Frady2018} (solid lines).}
    \label{fig:capacity}
\end{figure}

The probability that this inference finds the correct relationship can be predicted by the VSA capacity analysis \citep{Frady2018} (Fig. \ref{fig:capacity}). The probability is a function of the
%is can be computed by the $p_{corr}$ integral described in this paper and the 
signal-to-noise ratio, given in this case by $s^2 = N / R^2$. %Based on this theory, we can predict the accuracy of the analogy computation .

\subsubsection{Solving classification problems}
\label{mltasks}

%{\color{red}
%The previous subsection has demonstrated that the sparse binding operation can be used in cognitive reasons examples of VSA. 
%This subsection focuses on more applied use-cases of solving classification problems with VSA. 
Although VSAs originated as models for symbolic reasoning, \citet{intRVFL} have recently described the similarities between VSAs and randomly connected feed-forward neural networks \citep{RCNNSsurvey} for classification, known as Random Vector Functional Link (RVFL) \citep{RVFLorig} or Extreme Learning Machines (ELM) \citep{ELM06}. Specifically, 
%they have shown that 
RVFL/ELM can be expressed by VSA operations in the MAP VSA model \citep{intRVFL}.
%Due to the fact that most of VSA frameworks have similar functionally other frameworks can be used to solve classification problems as well. 
Leveraging these insights, we implemented a classification model using a VSA with sparse block-codes.

The model proposed in \citep{intRVFL} forms a dense distributed representation ${\mathbf x}$ of a set of features ${\mathbf a}$. 
Each feature is assigned a random ``key'' vector ${\bf \Phi}_i  \in \{\pm 1\}^N$. 
The collection of ``key'' vectors constitutes the  codebook $\mathbf{\Phi}$. 
However, in contrast to~(\ref{vsa_summation}) the set of features is represented differently. 
The proposed approach requires the mapping of a feature value $a_i$ to distributed representation ${\bf F}_i$ (``value'')  which preserve the similarity between nearby scalars. 
\citet{intRVFL} used thermometric encoding \citep{Scalarencoding} to create such similarity preserving distributed representations. 
The feature set is represented as the sum of ``key''-``value'' pairs using the binding operation:
\noindent
\begin{equation}
    {\mathbf x} =f_\kappa ( \sum_{i}^M {\bf \Phi}_i \odot {\bf F}_i ),
    \label{intRVFL}
\end{equation}
\noindent
where $f_\kappa$ denotes the clipping function which is used as a nonlinear activation function: 
\noindent
\begin{equation}
f_\kappa (x_i) = 
\begin{cases}
-\kappa & x_i \leq -\kappa \\
x_i & -\kappa < x_i < \kappa \\
\kappa & x_i \geq \kappa
\end{cases}
\label{eq:clipping}
\end{equation}
\noindent
The clipping function is characterized by the configurable threshold parameter $\kappa$ regulating nonlinear behavior of the neurons and limiting the range of activation values. 
\noindent

The predicted class $\hat{y}$ is read out from ${\mathbf x}$ using the trainable readout matrix as:
\begin{equation}
\hat{y} = \mbox{argmax} \textbf{W}^{\text{out}} {\mathbf x},
\label{intRVFL:readout}
\end{equation}
\noindent
where $\textbf{W}^{\text{out}}$ is obtained via the ridge regression applied to a training dataset.

For the purposes of using sparse block-codes, however, thermometric codes are non-sparse and their mean activity is variable across different values. Building on earlier efforts in the design of similarity-preserving sparse coding \citep{palm1994associative, palm2013neural},  
%encoding will not work in the case of sparse block-codes as the produced representations are neither sparse nor feature constant density. 
we design a similarity-preserving encoding scheme with sparse block-codes. 
%The encoding is extremely simply. 
In this scheme, the lowest signal level has all hot components in the first positions of each block. The second signal level is encoded by the same pattern except that the hot component of the first block is shifted to the second position. The third signal level is encoded by the code of the second level with the hot component of the second block shifted to the second position, and so on. 
%We continue in the same manner until obtaining the sparse block-code, which has its one-hot elements in the $N/K$th element of each block.
This feature encoding scheme can represent $N-K+1$ signal levels uniquely. 
\begin{figure}[tb]%[h]
\begin{center}
\begin{minipage}[h]{0.57\columnwidth}
\centering
\center{\includegraphics[width=1.0\columnwidth]{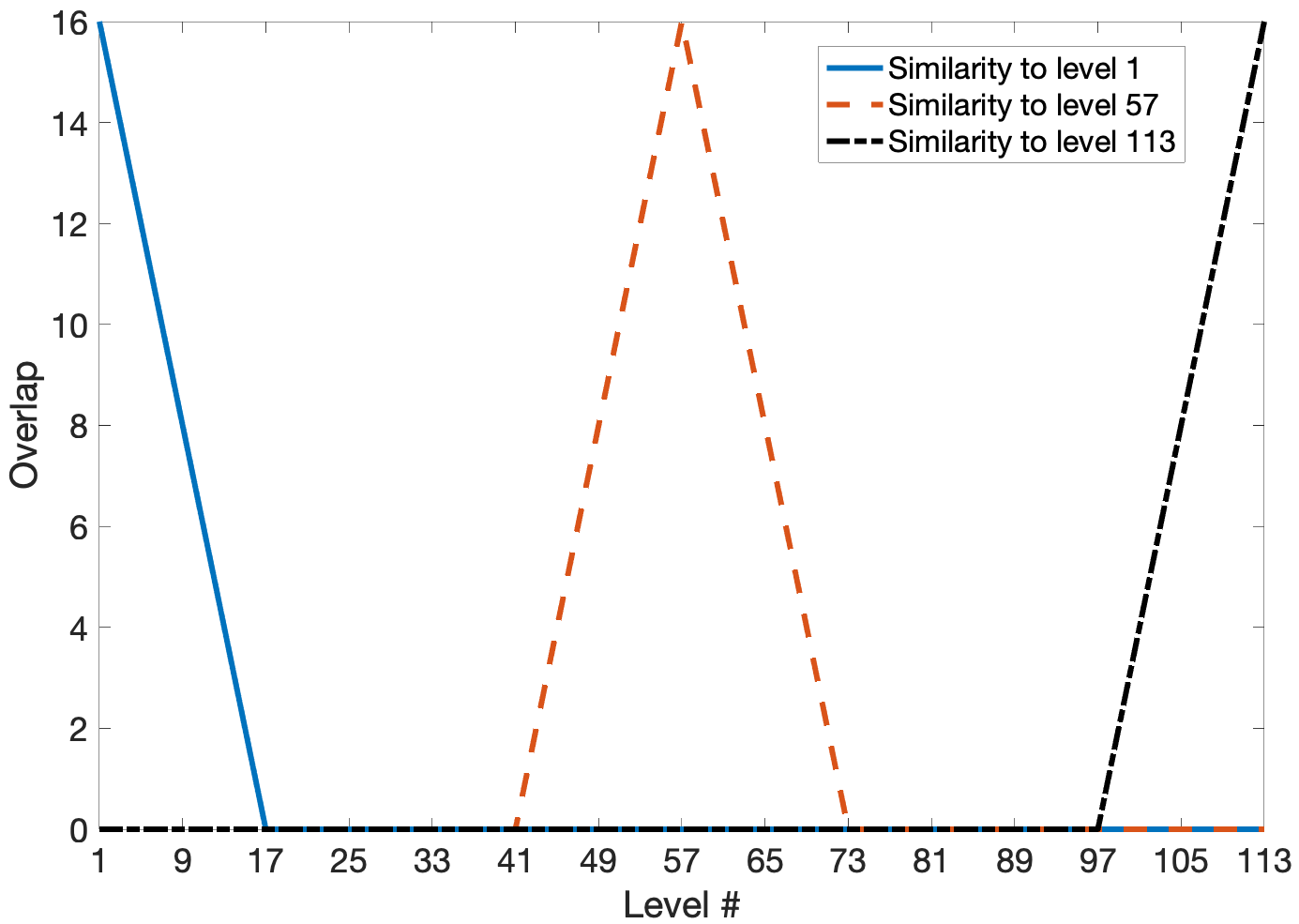}}\\A
\end{minipage}
\hfill
\begin{minipage}[h]{0.41\columnwidth}
\center{\includegraphics[width=1.0\columnwidth]{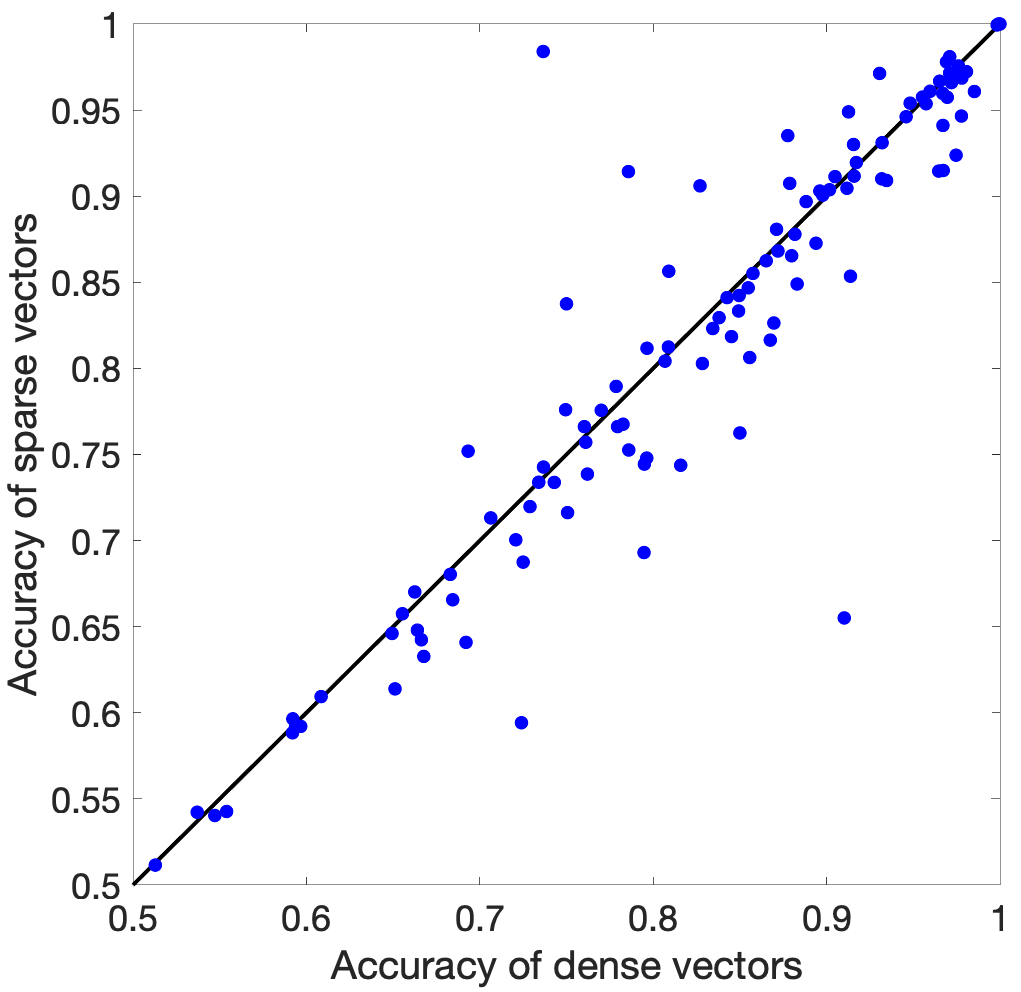}} \\B
\end{minipage}
\end{center}
\caption{
\textbf{Solving classification problems with sparse block-codes.} 
A. Similarity preserving representation of scalars with sparse block-codes: $K=16$, $N=128$. Similarity (overlap) between the representions ofthe levels $0, 64$ and the vectors representing other signal levels. 
B. Cross-validation accuracy of the VSA with dense distributed representations against the VSA with sparse distributed representations.
A point corresponds to a dataset.
}
\label{fig:perf:accuracy}
\end{figure}
The similarity between vectors drops of gradually as a function of distance (Fig.~\ref{fig:perf:accuracy}A).
Each pattern has the highest similarity with itself (overlap = $K$). For the range of distances between $1$ and $K$, 
%We also see that for $K$ levels, which are within the level of interest, 
the overlap decreases linearly until it reaches $0$ and then stays at this level for larger distances. 

The data vectors in a classification problem are encoded by the following steps. First, labels of the different features (data dimensions) are encoded by random sparse block-code vectors. Key-value pairs are then formed by binding feature labels with corresponding values, using the similarity preserving sparse block-code scheme described above.
%acting as value, the key-value pair sparse block-code is formed via the sparse binding operation. 
A data vector is then represented by the sum of all the key-value pairs. In essence, such a representation coincides is a protected sum (\ref{prot_sum}). In addition, we apply a clipping function to the resulting input.
%is then applied to the key-value pair sparse block-codes of all features. 
%The clipping function was used as a source of nonlinearity.

The described representation of the data can be computed in a sparse block-code VSA, the last step can be represented by the 
activation of a hidden layer with nonlinear neurons.
To perform classification, the hidden representation is pattern-matched to prototypes of the different classes. To optimize this pattern matching, in cases where the prototypes are correlated, we train a perceptron network with ridge regression, similar as previously proposed for in a sequence memory with VSAs \citep{Frady2018}.  
%multiplied by the readout matrix to form a vector of scores encoding the strength of membership to the different classes.  The class with the highest score is considered to be the best prediction. The readout matrix (real-valued and dense) is obtained by ridge regression on the training datasets.

Interestingly, the cross-validated accuracies for VSAs with sparse block and dense representations \citep{intRVFL} on $121$ real-world classification datasets are quite similar (Fig.~\ref{fig:perf:accuracy}B), with a correlation coefficient of $0.88$ and both reaching average accuracy of $0.80$.
The $121$ datasets from the UCI Machine Learning Repository \citep{Dua:2019} have been initially analyzed in a large-scale comparison study of different classifiers \citep{HundredsClassifiers}. The only preprocessing step we introduced was to normalize features in the range $[0, 1]$ and quantize the values into $N-K+1$ levels.
The hyperparameters for both dense and sparse models were optimized through grid search over $N$ 
(for dense representations $N$ varied in the range $[50, 1500]$ with step $50$), $\lambda$ (ridge regression regularization parameter; varied in the range $2^{[-10, 5]}$ with step $1$), and $\kappa$ (varied between $\{1,3,7,15\}$).
%$\kappa$ is a parameter of the the clipping function, which is used as a nonlinear activation function for the hidden layer. It limits the activation values of neurons in the range $[-\kappa, \kappa]$. 
The search additionally considered $K$ for sparse block-codes ($K/N$ varied in the range $2^{[2, 5]}$ with step $1$ while $K$ varied in the range $2^{[4, 7]}$ with step $1$).
%The obtained optimal hyperparameters were used to estimate the cross-validation accuracy on all datasets as depicted in Fig.~\ref{fig:perf:accuracy}B. 
%First, it is important to note that as expected the between the obtained results was high ($0.88$). 
%Moreover, the average accuracy for both approaches was approximately .

Importantly, the average number of neurons used by both approaches was also comparable: about $500$ for sparse block-codes  and about $750$ for dense representations. Thus, we conclude that sparse block-codes can be used as substitutes of dense representations for practical problems such as classifications tasks. 
%}

\section{Discussion}
\label{sec:discu}
In this paper we investigated methods for variable binding for symbolic reasoning with sparse distributed representations. The motivation for this study was two-fold. First, we believe that such methods of variable binding could be key for combining the universal reasoning properties of vector symbolic computing \citep{Gayler2003} with advantages of neural networks.  Second, these methods will enable implementations of symbolic reasoning that can leverage efficient sparse Hebbian associative memories \citep{willshaw1969non, palm1980associative,KnoblauchPalm2019} and low-power neuromorphic hardware \citep{Davies2018}. 

\subsection{Theoretical Results}
Using the framework of compressed sensing, we investigated a setting in which there is a unique equivalence between sparse feature vectors and dense random vectors.  
%We reviewed the existing models for vector symbolic reasoning with dense vectors and employed the compressed sensing framework to understand their properties and relationships from a theoretical point of view. 
We find that: 
\begin{itemize}
\item[i)] 
With this setting, CS inference outperforms the classical VSA readout of set representations.

\item[ii)] Classical vector symbolic binding between dense vectors with the Hadamard product \citep{Plate2003, Gayler1998, Kanerva2009} is under certain conditions mathematically equivalent to tensor product binding \citep{Smolensky1990} of the corresponding sparse vectors. 

\item[iii)] For representing sets of objects, vector addition of dense vectors (\ref{eqn:addition}) is equivalent to addition of the corresponding sparse vectors. %However, the original sparse vectors cannot be decoded from the sum representation because a binding problem occurs, which destroys the integrity of individual sparse vectors.

\item[iv)] The protected sum of dense vectors (\ref{eqn:set}) is equivalent to the concatenation of the sparse vectors. 
%Individual sparse vectors can be trivially decoded from their concatenation. However, to infer the sparse concatenation vector from the dense protected sum becomes more challenging with increasing number of concatenations, as described in \citep{Frady2018}.

%\item[iv] In classical VSAs, sets of sets cannot be represented by just applying vector summation twice because the set information represented by the first sum operation is lost. Introducing variable binding between the two summations yields a protected set representation which preserves the set information.

\item[v)] The dimensionality preserving operations between dense vectors for variable binding and protected set representations mathematically correspond to operations between sparse vectors, tensor product and vector concatenation, which are not dimensionality preserving. 
\end{itemize}

\subsection{Experimental Results}
%As an experimental demonstration of the value of applying CS theory to VSAs, we showed that  
%The most important theoretical result, however, is that the CS equivalence between dense and sparse representations does not directly induce sparsity-preserving operations for variable binding and protected set representation. 
%The CS equivalence further shows that VSA binding operations can be reduced to a projection of the tensor-product. 
Our theory result v) implies that in order to construct dimensionality and sparsity-preserving variable binding between sparse vectors, an additional reduction step is required for mapping the outer product to a sparse vector. 
%The tensor-product is the basis for binding in TPR models conceived by \citet{Smolensky1990}.  But tensor product binding does not fit into the VSA scheme because the dimension of compound representations depends on the complexity of the composition and can become very large. 
Existing reduction schemes of the outer product proposed in the literature, circular convolution \citep{Plate2003} and vector-derived transformation binding \citep{gosmann2019}, are not sparsity-preserving when applied to sparse vectors.
%The existing schemes for reducing the tensor product can be made sparsity-preserving by introducing a threshold nonlinearity. 

For binding pairs of general $K$-sparse vectors, we designed a strategy of sub-sampling from the outer-product with additional thresholding to maintain sparsity. Such a computation can be implemented in neural circuitry where dendrites of neurons detect firing coincidences between pairs of input neurons. The necessary connection density increases with sparsity of the code vectors. Still, the computational complexity is order $N$ when $K=\beta N$, which favorably compares to other binding operations which can have order of $N^2$ or $N \log N$. However, 
%with general sparse random vectors, 
the sampling in the circuit always misses components of the tensor product, making the unbinding operation lossy.   

Another direction we investigated extends previous work \citep{Laiho2015} developing VSAs for sparse representations of restricted type, sparse block-codes. We propose block-wise circular convolution as a variable binding method which is sparsity and dimensionality preserving.
%For such codes, Laiho et al proposed  
Interestingly, for sparse block-codes, the unbinding given the reduced tensor and one of the factors is lossless. As our experiments show, it has the desired properties required for VSA manipulations, outperforming the other methods. Independent other work has proposed efficient Hebbian associative memory models \citep{Potts, gripon2011,KnoblauchPalm2019} that could be applied for cleanup steps required in VSAs with block-codes.

VSAs with block-codes are demonstrated in two applications. In a symbolic reasoning application we show that the accuracy as a function of the dimension of sparse block-codes reaches the full performance of dense VSAs and can be described by the same theory \citep{Frady2018}. On $121$ classification datasets from the UCI Machine Learning Repository we show that the block-code VSA reaches the same performance as dense VSAs \citep{intRVFL}.
Moreover, the average accuracy of $0.80$ of VSAs models is comparable to the state-of-the-art performance of $0.82$ achieved by Random Forest \citep{HundredsClassifiers}.

%In contrast, sparse vectors resulting from block-codes  guarantee that the binding can be represented in a sparse vector without loss. So far, sparse block-codes provide the best results in our experiments based on how much information is retained when binding and unbinding. Currently we are exploring if there are other other types of structured vectors with similar desirable properties.

\subsection{Relationship to earlier work}
\citet{Rachkovskij2001, Rachkovskij2001a} were to our knowledge the first to propose similarity- and sparsity-preserving variable binding. For binary representations they proposed methods that involve component-wise Boolean operations and deletion (thinning) based on random permutations. These methods of variable binding are also lossy, similar to our method of SPTP.

The variable binding with block-codes, which our experiments identify as the best, can be done with real-valued binary or complex-valued phasor block codes. For binary block-codes our binding method is the same as in \citep{Laiho2015}, who demonstrated it in a task processing symbolic sequences. %This framework can also be extended to the complex domain. 
For protecting individual elements in a sum representation, they use random permutations between blocks, rather than variable binding as we do in section~\ref{sec:protsum}.

\subsection{Implications for neural networks and machine learning}
In the deep network literature, concatenation is often used in neural network models as a variable binding operation \citep{Soll2019}. %\hl{\textbf{(CITE)}}. 
However, our result iv) suggests that concatenation is fundamentally different from a binding operation. This might be a reason why deep learning methods have limited capabilities to represent and manipulate data structures \citep{MarcusAI2020}. %This could explain some of their seemingly fundamental weaknesses, such as the inability to generalize and extrapolate learned knowledge, and the need of large amounts of training data \textbf{(CITE)}. [Recent smolensky product paper, deep learning nets with multiplicative attention windows, deep nets with hadamard products--trevor darrel]

Several recent studies have applied VSAs to classification problems~\citep{HDClassGe, HDGestureIEEE}. Here we demonstrated classification in a block-code VSA. 
%to create randomly connected feed-forward neural networks. 
The block-code VSA exhibited the same average classification accuracy as earlier VSA solutions with dense codes. This result suggests that sparse block-code VSAs can be a promising basis for developing classification algorithms for low-power 
%means that there are classes of problems that can be successfully solved by networks with sparse activity patterns, which in turn is favorable for efficient implementations in 
neuromorphic hardware platforms \citep{Davies2018}.

\subsection{Implications for neuroscience}
We have investigated variable binding operations between sparse patters regarding their computational properties in symbolic reasoning. It is interesting that this form of variable binding requires multiplication or coincidence detection, computations which can be implemented by active dendritic mechanisms of biological neurons \citep{larkum2008synaptic}. Although this computation is beyond the capabilities of standard neural networks, it can be implemented with formal models of neurons, such as sigma-pi neurons \citep{mel1990sigma}.

We found that the most efficient form of variable binding with sparse vectors relies on block-code structure.
Although block-codes were engineered independent of neurobiology, they 
compatible with some experimental observations, such as divisive normalization \citep{heeger1992normalization}, and functional modularity. Specifically, in sensory cortices of carnivores neurons within small cortical columns \citep{mountcastle1957modality} respond to the same stimulus features, such as the orientation of local edges in the image \citep{hubel1963,hubel1977ferrier}. Further, groups of nearby orientation columns form so-called macro columns, tiling all possible edge orientations at a specific image location \citep{hubel1974sequence,swindale1990cerebral}. A macro column may correspond to a block in a block-code.
%, with the active column corresponding to the active element within the block.

While binary block-codes are not biologically plausible, 
complex-valued block-codes in which active elements are complex phasors with unit magnitude, can be represented as timing patterns in networks of spiking neurons \citep{frady2019}.  Further, it seems possible to extend LCC binding to soft block-codes, in which localized bumps with graded neural activities represented by spike rate, e.g. \citep{ben1995theory}.  
   %However, a full description of biophysical mechanisms for implementing binding with spikes is yet to be described.
 
\subsection{Future directions}
%Our results open up a number of future directions in research. 
One important future direction is to investigate how to combine the advantages of VSA and traditional neural networks to build more powerful tools for artificial intelligence. The challenge is how to design neural networks for learning sparse representations that can be processed in sparse VSAs. 
%Fortunately, there are efficient associative memory models for block-codes \citep{KnoblauchPalm2019} that can be seamlessly processed in sparse VSAs. 
Such combined systems could potentially overcome some of the severe limitations of current neural networks, such as the demand of large amounts of data, limited abilities to generalize learned knowledge, etc. 

Another interesting research direction is to design VSAs operating with spatio-temporal spike patterns that can be implemented in neuromorphic hardware, potentially also making use spike timing and efficient associative memory for spike timing patterns \citep{frady2019}. 

Further, it will be interesting to study how binding in sparse VSAs can be used to form similarity-preserving sparse codes \citep{palm1994associative,palm2013neural} for continuous manifolds. 
For example, binding can be used to create index patterns for representing locations in space, which could be useful for navigation in normative modeling of hippocampus \citep{FradySommer2020}. 
%  from similarity-preserving coding schemes for a one-dimensional manifold, for example, for building models for hippocampus \citep{FradySommer2020}, navigation systems, etc.

\section*{Acknowledgement}
The authors thank Charles Garfinkle and other members of the Redwood Center for Theoretical Neuroscience for stimulating discussions. DK is supported by the European Union’s Horizon 2020 research and innovation programme under the Marie Skłodowska-Curie Individual Fellowship grant agreement No. 839179, and DARPA's VIP program under Super-HD project. FTS is supported by NIH R01-EB026955. 

\appendices
%\section{Appendix}

\section{Relations between different variable binding operations}
\label{sec:classical_vsa_binding}

\subsection{VSA binding, a subsampling of TRP}
The dimensionality-preserving binding operations in VSAs can be expressed as a sampling of the tensor product matrix ${\mathbf x} \; {\mathbf y}^{\top}$ into a vector:
\begin{equation}
    {\mathbf x} \circ {\mathbf y} = 
    %w\left({\mathbf x} \; {\mathbf y}^{\top}\right) = 
    \sum_{ij} W^l_{ij} x_i y_j 
    %= \mathbf{W} \left({\mathbf x} \; {\mathbf y}^{\top}\right)
    \label{tensor_product_projection}
\end{equation}
where 
%the linear projection $\mathbf{W}: \mathbb{R}^{N \times N} \to \mathbb{R}^{N}$ is defined by 
the binary third-order tensor $\mathbf{W} \in \{0,1\}^{N \times N \times N}$ determines what elements of the outer-product are sampled. %Different VSAs use different types of sampling tensors. 
For the Hadamard product in the MAP VSA, the sampling tensor just copies the diagonal elements of the tensor matrix into a vector, using a sampling tensor $W^l_{ij} = \delta(i, j)\delta(i, l)$. Here $\delta(i,j)$ is the Kronecker symbol.
Conversely, in circular convolution 
%\begin{equation}
%    {\mathbf x} \ast {\mathbf y} = w_{\ast}\left( {\mathbf x} \; %{\mathbf y}^{\top} \right)
%    \label{circonvapx}
%\end{equation}
the sampling involves summing the diagonals of the outer-product matrix:
\begin{equation}
    W^l_{ij} = \delta((i+l-1)_{mod \, n}, j)
    \label{circonv_sampling}
\end{equation}
%Here will explore whether a similar binding operation can be made sparsity-preserving by adding a threshold operation. 
%\subsection{Generalization of the Fourier convolution theorem}
%\label{sec:appendix-genfourier}

For neurally implementing a binding operation, like (\ref{sptp}), a low fan-in is essential. 
The {\it fan-in} is the number of nonzero elements in the tensor feeding the coincidences between the input vectors to an output neuron $\alpha=\alpha(l)=\sum_{ij} W^l_{ij}$. For circular convolution the fan-in is $\alpha_{CCB}=N$. For VDTB binding the fan-in is $\alpha_{VDTB}=\sqrt{N}$. When applied to a pair of sparse vectors, circular convolution and VDTB binding are not sparsity-preserving. In the next section we analyze the properties of the sampling tensor (\ref{tensor_product_projection}) required to make (\ref{sptp}) a sparsity-preserving binding operation for general $K$-sparse vectors with optimal properties.

\subsection{Generalizing the Fourier Convolution Theorem}

There is a direct relation between circular convolution binding and the Hadamard product binding through the Fourier convolution theorem. The Fourier transform is a (non-random) linear transform that previously has been proposed in holography for generating (dense) distributed representations from data features. 
%Interestingly, the relationship between sparse and dense variable binding (\ref{bind}) is a generalization of the Fourier convolution theorem. 
%For the circular convolution (\ref{circonv}) 
The {\it Fourier convolution theorem} states:
\begin{equation}
{\cal F}({\bf a}) \odot {\cal F}({\bf b}) = {\cal F}({\bf a} \ast {\bf b})
\label{fct}
\end{equation}
where ${\cal F}(z) := (1/n)\sum_{m=0}^{n-1} \Phi^F_{-km} z_m$ with $\Phi^F_{km} := e^{j\frac{2\pi km}{n}}$ is the discrete Fourier transform.
With (\ref{tensor_product_projection}) and (\ref{circonv_sampling}), 
%one can rewrite the RHS of (\ref{fct})
%\begin{equation}
%{\cal F}({\bf a} \ast {\bf b}) = {\cal F}(w_{\ast} ({\bf a} \; {\bf b}^{\top}))
%\label{fctlh}
%\end{equation}
%with $w_{\ast}()$ the linear projection of circular convolution (\ref{circonvapx}). Thus, 
the Fourier convolution theorem establishes a relationship between the outer product of two vectors and the Hadamard product of their Fourier transforms. 
%The particular form of this relationship rests on the following property of the Fourier matrix:  
%$\Phi^F_{ki} \Phi^F_{kl} = \Phi^F_{k(i+l)} \;\; \forall i,k,l$.
Replacing the Fourier transform by 
%In the case of the Hadamard product of two vectors that are transformed by 
two CS random sampling matrices, 
%two different linear transforms ${\cal G}$ and ${\cal H}$, 
the Fourier convolution theorem generalizes to:
\begin{equation}
\mathbf{\Phi}{\bf a} \odot \mathbf{\Psi}{\bf b} = {\cal J}({\bf a} \; {\bf b}^{\top})
\label{gfct}
\end{equation}
This equation coincides with (\ref{bind}), describing the relationship between the Hadamard product of dense vectors to the outer product of the corresponding sparse vectors. The linear projection ${\cal J}$ is formed from the CS sampling matrices $\mathbf{\Phi}$ and $\mathbf{\Psi}$. Note, that there is no general invertibility of ${\cal J}$, unlike in the Fourier convolution theorem. The outer product of the sparse vectors can be uniquely inferred from the Hadamard product of the dense vectors under certain conditions of sparsity, dimensionality, and properties of the sampling matrices, as discussed in Sect.~\ref{hadamard_tensor}.   

%in which (\ref{fctlh}) can be rewritten as a convolution operation yielding the RHS of (\ref{fct}). Note that the wrap-around symmetry is a particular property of the Fourier transform, however, (ref{genconv}) also applies to transforms that do not fulfill this property. 
%Note, however, that the Fourier transform is only fully distributed (i.e., all frequencies present) for feature representations that are extremely sparse and noiseless. 

\section{Sparsity-preserving binding for $K$-sparse vectors}

\subsection{Analysis of required fan-in}
\label{sec:fan-in}
%Here we analyze  our sparsity-preserving binding method for general sparse vectors (\ref{sptp}).
%Similar to CCB and VDTB, we propose a sampling matrix in TPS for subsampling the tensor product. Subsampling in TPS is denoted by the operator $\underline{\times}$. The fan-in in the subsampling matrix, $\alpha$, is the same for all downstream neurons. 
We examine the required fan-in in (\ref{sptp}) for two types of random sampling tensors for sparsity-preserving binding (Fig.~\ref{model}). One type in which the tensor product of the input vectors is sampled entirely randomly with a fixed constant fan-in per downstream neuron. And another type, in which the tensor is sampled along its diagonals, similar to circular convolution, but the diagonals truncated by a fixed sized fan-in of the downstream neurons.
%where each neuron samples along a diagonal line in the outer product. In structured TPS, these diagonal samples are truncated to the size of the fan-in.

First, we determine the minimal fan-in that still provides signals at downstream neurons so that a threshold operation can reliably produce a $K$-sparse vector. 
The dendritic sums in (\ref{sptp}) are approximately distributed by the following Binomial distribution:
\begin{equation}
    p(d_l=r) = {\alpha \choose r} \left(\frac{K^2}{N^2}\right)^r \left(1- \frac{K^2}{N^2}\right)^{\alpha -r}
    \label{dendsumdistri}
\end{equation}
We require that the fan-in is large enough so that the expected number of downstream neurons with $d_l=0$ is smaller than the number of silent neurons in the $K$-sparse result vector: $N p(d_l=0) < N-K$. To satisfy this condition, a lower bound $\alpha^*$ to the minimal fan-in is computed, which ensures the equality between the two numbers and translates into the condition that for a threshold of $\theta = 1$ the patterns sparsity is preserved:
\begin{equation}
    % what is the \eoe{=} supposed to mean?
    p(d_l=0) \eoe{=} 1 - \frac{K}{N}
    \label{sparsitypreservation_teq1}
\end{equation}
Inserting (\ref{dendsumdistri}) into this condition we obtain after some algebra: 
\begin{equation}
    \alpha > \alpha^* = \frac{\ln\left(1-\frac{K}{N}\right)}{\ln\left(1-\frac{K^2}{N^2}\right)}
    \label{alphstar}
\end{equation}
For sparse patterns, (\ref{alphstar}) becomes simply $\alpha^* = N/K$. 

\begin{figure}[t]
    \centering
    \includegraphics[width=0.35\textwidth]{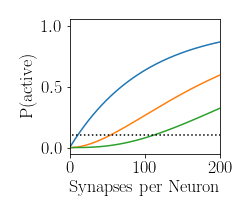}
    \caption{\textbf{Fan-in requirements of SPTP.} The fan-in for each neuron can be determined from the binomial distribution. Higher thresholds require more fan-in ($\theta=[1,2,3]$). The fan-in is determined where each colored line crosses the sparsity level (black line shows 10\% sparsity).}
    \label{fig:tps}
\end{figure}

Note, that setting the fan-in exactly to the lower bound $\alpha^*$ should result in patterns of dendritic activity in the population of downstream neurons which are approximately $K$-sparse, without any thresholding necessary. One can generalize condition (\ref{sparsitypreservation_teq1}) to an arbitrary threshold:
\begin{align}
\begin{split}
    \sum_{i=0}^{\theta -1} p(d_l=i) &= (\alpha - \theta +1) {\alpha \choose \theta-1} \int_0^{1-\frac{K^2}{N^2}} t^{\alpha-\theta}(1-t)^{\theta-1} dt \\
    &\eoe{=} 1 - \frac{K}{N}
    \label{sparsitypreservation}
\end{split}
\end{align}
Unfortunately, the variable $\alpha$ cannot be analytically resolved from the exact condition (\ref{sparsitypreservation}). However, it is straight-forward to compute $\alpha$ numerically (Fig.~\ref{fig:tps}A).
% Replacing the exact calculation with an approximation based on the Chernoff bound $p(d_l \leq \theta') \leq e^{-2\frac{(\alpha q-\theta')^2}{\alpha}} \eoe{=} 1 - \frac{k}{n}$ with $\theta':= \theta-1$, $q := \frac{k^2}{n^2}$ and $s:=-\ln(1-\frac{k}{n})$, one obtains for $\alpha^*$:
% \begin{equation}
%     \alpha_{1,2} = \frac{1}{q} \left(\theta' + \frac{s}{4q}\right) \pm \sqrt{\frac{s}{2 q^3}\left(\theta' +\frac{s}{8q}\right)} 
%     \label{alphachernoff}
% \end{equation}
% For $\theta'=0$ and small $q$ (\ref{alphachernoff}) yields the result $\alpha^* = - \ln(1-\frac{k}{n})/(2 q) = \frac{n}{2 k}$, which is off from the correct result by a factor of two.

%The situation is somewhat easier because the tensor is constructed deterministically to exclusively tile the elements to the outer product so that there is no overlap between the inputs of different downstream neurons. 
%For CCB each neuron receives $n$ neurons and the probability that a neuron receives no input is :$p(d_l=0) = (1-k^2/n^2)$  In VDTB the number weights in the network is smaller, only $O(\sqrt{n^3})$. As the connectivity pattern in CCB and VDTB is chosen deterministically so to 

\subsection{Symmetry for optimizing unbinding performance}
\label{sec:appendix_symmetry}
Another crucial question is what symmetry of the $\mathbf{W}$ tensor best enables the inversibility of the sparsity-preserving binding operation (\ref{sptp}).
%\begin{equation}
 %   a_i = H\left(\sum_{jl} W^i_{jl} b_j c_l -\theta \right)
%\end{equation}
Chaining a binding and unbinding step, one obtains the following self-consistency condition:
\begin{equation}
    a_i = H\left(\sum_{j,l} W^i_{jl} b_j H\left(\sum_{i',j'} W^l_{i' j'} a_{i'} b_{j'} -\theta \right) -\theta \right)\label{selfconsist}
\end{equation}
which should hold for arbitrary $K$-sparse vectors $\mathbf{a}$ and $\mathbf{b}$. 
The self-consistency condition can be approximately substituted by maximizing the objective function ${\cal L}:= \sum_i a_i (d_i-\theta)$. Replacing also the inner nonlinearity by its argument we obtain:
\begin{align}
\begin{split}
    {\cal L}(W^i_{jl}; \mathbf{a}, \mathbf{b},\theta) = & \sum_{i,j, i', j'} W^i_{ij} W^l_{i' j'} a_i a_{i'} b_j b_{j'} \\
             & - \theta \left( \sum_{ijl} W^i_{jl} a_i b_j + K \right) 
\label{objective}
\end{split}
\end{align}
The quantity (\ref{objective}) should be high for any vectors $\mathbf{a}, \mathbf{b}$. Thus it is only the expectation (over all vectors $\mathbf{a}$ and $\mathbf{b}$) of first term that can be consistently increased by changing the structure of the sampling tensor. The biggest increase is achieved by making sure that terms with $(a_i)^2 (b_j)^2$ are zeroed out with probability $1-\alpha$ rather than with $1-\alpha^2$, which can be accomplished by introducing the following symmetry into the sampling tensor:
\begin{equation}
    W^i_{jl} = W^l_{ij}
\end{equation}

For sparse complex vectors one can define a binding operation with the nonlinearity $f_{\Theta}(x) = \frac{x}{|x|} H(|x|-\Theta)$ taken from \citep{frady2019}:
\begin{equation}
    {\mathbf a} \circ {\mathbf b} = f_{\Theta}\left(\sum_{jl} W^l_{ij} a_i b_j\right)
\end{equation}
where the sampling tensor is binary random or a random phasor tensor.
The corresponding selfconsistency condition is then:
\begin{align}
    \begin{split}
    a_i &= f_{\Theta}\left(\sum_{j,l} W^i_{jl} \bar{b}_j \; f_{\Theta}\left(\sum_{i',j'} W^l_{i' j'} a_{i'}  b_{j'}\right)\right) \\
    &\simeq f_{\Theta}\left(\sum_{j,l,i',j'} W^i_{jl} W^l_{i' j'} a_{i'}  b_{j'} \bar{b}_j \right)
    \label{selfconsistcomplex} 
\end{split}
\end{align}

In (\ref{selfconsistcomplex}) one can notice that the signal is maximized if the tensor fulfills the following symmetry:
\begin{equation}
    W^i_{jl} = \bar{W}^l_{ij}
    \label{herm}
\end{equation}
Even with condition (\ref{herm}) the unbinding will be noisy and cleanup through an additional associative network will be required.

% --------- 
 
% Hinton, G. E. (1990). Mapping part-whole hierarchies into connectionist networks. Artificial Intelligence, 46, 47-76. 

%\bibliographystyle{abbrvnat} 
\bibliographystyle{apalike}

\bibliography{capacity,hdspike,additionalrefs}

\end{document}